\documentclass[final]{cvpr}

\usepackage{nopageno}

\usepackage{times}
\usepackage{epsfig}
\usepackage{graphicx}
\usepackage{amsmath}
\usepackage{amssymb}
\usepackage{bm}
\usepackage{subfig}
\usepackage{wrapfig}
\usepackage[outline]{contour}
\usepackage{multirow}
\usepackage{enumitem}
\usepackage{microtype}
\usepackage{booktabs}
\usepackage{amsfonts}
\usepackage{nicefrac}

\DeclareMathOperator*{\argmax}{argmax}
\usepackage[pagebackref=true,breaklinks=true,colorlinks,bookmarks=false]{hyperref}

\def\cvprPaperID{6533} % *** Enter the CVPR Paper ID here
\def\confYear{CVPR 2021}

\begin{document}

%%%%%%%%% TITLE
\title{Hierarchical and Partially Observable Goal-driven Policy Learning \\ with Goals Relational Graph}

\author{Xin Ye and Yezhou Yang\\
Active Perception Group, School of Computing, Informatics, and Decision Systems Engineering, \\ Arizona State University, Tempe, USA\\
% Institution1 address\\
{\tt\small \{xinye1, yz.yang\}@asu.edu}
% For a paper whose authors are all at the same institution,
% omit the following lines up until the closing ``}''.
% Additional authors and addresses can be added with ``\and'',
% just like the second author.
% To save space, use either the email address or home page, not both
% \and
% Second Author\\
% Institution2\\
% First line of institution2 address\\
% {\tt\small secondauthor@i2.org}
}

\maketitle

%%%%%%%%% ABSTRACT
\begin{abstract}
   We present a novel two-layer hierarchical reinforcement learning approach equipped with a Goals Relational Graph (GRG) for tackling the partially observable goal-driven task, such as goal-driven visual navigation. Our GRG captures the underlying relations of all goals in the goal space through a Dirichlet-categorical process that facilitates: 1) the high-level network raising a sub-goal towards achieving a designated final goal; 2) the low-level network towards an optimal policy; and 3) the overall system generalizing unseen environments and goals. We evaluate our approach with two settings of partially observable goal-driven tasks --- a grid-world domain and a robotic object search task. Our experimental results show that our approach exhibits superior generalization performance on both unseen environments and new goals~\footnote{Codes and models are available at \url{https://github.com/Xin-Ye-1/HRL-GRG}.}.
\end{abstract}

%%%%%%%%% BODY TEXT
\section{Introduction}
\begin{figure*}[t]
\centering
\includegraphics[width=0.8\textwidth]{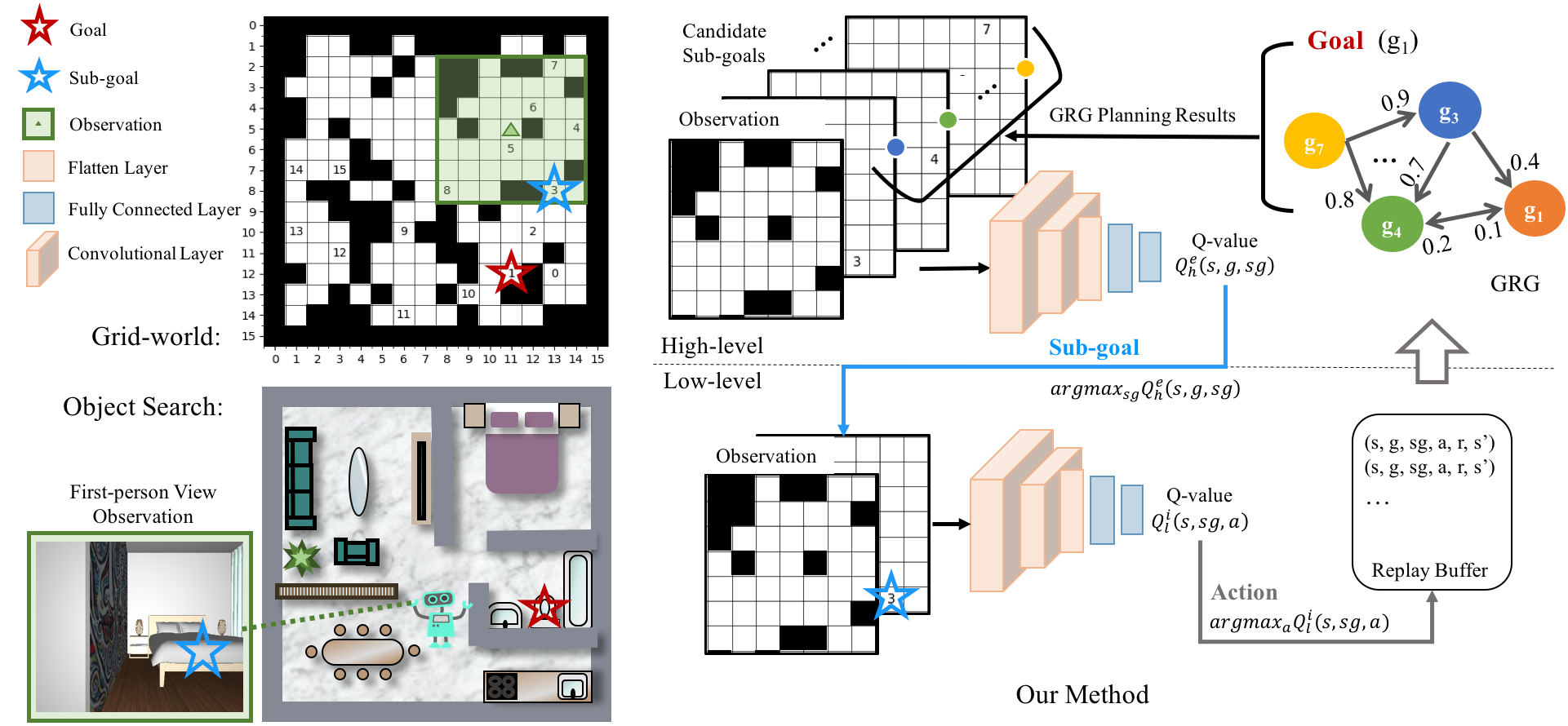}
\caption{Illustrations of the grid-world domain and the robotic object search task (left), and an overview of our method (right).}
\label{fig:overview}
\end{figure*}

Goal-driven visual navigation defines a task where an intelligent agent (with an on-board camera) is expected to take reasonable steps to navigate to a user-specified goal in an unknown environment (see Figure~\ref{fig:overview} left). It is a fundamental yet essential capability for an agent and could serve as an enabling step for other tasks, such as Embodied Question Answering \cite{das2018neural} and Vision-and-Language Navigation \cite{Anderson_2018_CVPR}. Goal-driven visual navigation could be formulated as a partially observable goal-driven task.
In this paper, we present a novel Hierarchical Reinforcement Learning approach with a Goals Relational Graph formulation (HRL-GRG) tackling it.  
%the partially observable goal-driven task.
Formally, a partially observable goal-driven task yields a $10$-tuple $<S, A, T, G, R, \Omega, O, G_d, \Phi, \gamma>$, in which $S$ is a set of states, $A$ is a set of actions, $T: S\times A \times S \to [0,1]$ is a state transition probability function, $G \subseteq S$ is a set of goal states, $R: S\times A \times S \times G \to \mathbb{R}$ is a reward function, $\Omega$ is a set of observations that are determined by conditional observation probability $O: S \times \Omega \to [0,1]$. Similarly, $G_d$ is a set of goal descriptions that describe observations with goal recognition probability $\Phi: \Omega \times G_d \to [0,1]$. In particular, $g$ is the goal state of the corresponding goal description $g_d$ iff $\Phi(\argmax_{\omega}O(g, \omega), g_d)$ is larger than a pre-defined threshold. $\gamma \in (0, 1]$ is a discount factor. The objective of a partially observable goal-driven task is to maximize the expected discounted cumulative rewards $\mathbb{E}[\sum_t^\infty \gamma^t r_{t+1}(s_t,a_t,s_{t+1},g)|s_t,g]$ by learning an optimal action policy to select an action $a_t$ at the state $s_t$ given the observation $o_t$ and the goal description $g_d$.

Classic RL methodology optimizes an agent's decision-making action policy in a given environment \cite{sutton2018reinforcement}. To make RL towards real-world applicable, equipped with deep neural networks, Deep Reinforcement Learning (DRL) \cite{mnih2015human} algorithms are able to directly take the high dimensional sensory inputs as states $S$ and learn the optimal action policy that generalizes across various states.
% learn the optimal action policy directly from high dimensional sensory inputs and generalize across various inputs.
However, the applicability of most advanced RL algorithms is still limited to domains with fully observed state space $S$ and/or fixed goal states $G$, which is not the case in reality \cite{mnih2015human,mnih2016asynchronous,lillicrap2015continuous,schulman2017proximal,haarnoja2018soft}.

% situations about real world applications
For real-world applications like visual navigation, an agent's sensory inputs capture the local information of its surrounding environments (a partially observable state space). Additionally, the real-world applications could be subject to goal changes, requiring a system to be goal-adaptive. 
Therefore, a real-world application can be formulated as a partially observable goal-driven task, that is different from a fully observable goal-driven task \cite{mnih2015human,mnih2016asynchronous,haarnoja2018soft,nasiriany2019planning} or a partially observable task \cite{igl2018deep,lee2019stochastic,han2019variational}. It requires the agent to be capable of inferring its state in the augmented state space $S \times G$.
Namely, the agent should take actions based on its current relative states with respect to the goal states, which can only be estimated from its sensory observations $\Omega$ and the goal descriptions $G_d$. 
% This is challenging due to lack of information, namely: 1) the goal can hardly be inferred from the agent's partially observable sensory inputs, and 2) specifying the goal with complete information is also intractable.
This is challenging due to 1) the large augmented state space, 2) the different modalities that the observations $\Omega$ and the goal descriptions $G_d$ could have. For example, while RGB images are usually taken as the observations, semantic labels are more efficient in describing task goals \cite{batra2020objectnav}.

% Bridging the real-world needs with DRL, our HRL-GRG decomposes the partially observable goal-driven task into two sub-tasks: 1) a high-level sub-goal selection task, and 2) a low-level fully observable goal-driven task. In particular, the high-level layer selects a sub-goal $sg \in G_d$ that is observable in the current sensory input $o$, i.e. $\Phi(o, sg) > 0$, and could also contribute to achieving the designated final goal $g \in G_d$. Our model is the first to incorporate the novel Goals Relational Graph (GRG), which is designed to learn goal relations from the training data through a Dirichlet-categorical process \cite{tu2014dirichlet} dynamically. In such a way, our model enhances the sub-goal selection efficiency of the high-level layer and improves the generalization ability across environments and goals. The objective of the low-level layer is to achieve the observable sub-goal, yielding a well-studied fully observable task   \cite{mnih2015human,mnih2016asynchronous,haarnoja2018soft}.

To address the challenges, our HRL-GRG incorporates a novel Goals Relational Graph (GRG), which is designed to learn goal relations from the training data through a Dirichlet-categorical process \cite{tu2014dirichlet} dynamically. In such a way, our model estimates the agent's states in terms of the learned relations between sub-goals that are visible in the agent's current observations and the designated final goal. Furthermore, our HRL-GRG decomposes the partially observable goal-driven task into two sub-tasks: 1) a high-level sub-goal selection task, and 2) a low-level fully observable goal-driven task. Specifically, the high-level layer selects a sub-goal $sg \in G_d$ that is observable in the current sensory input $o$, i.e. $\Phi(o, sg) > 0$, and could also contribute to achieving the designated final goal $g \in G_d$. The objective of the low-level layer is to achieve the observable sub-goal, yielding a well-studied fully observable task   \cite{mnih2015human,mnih2016asynchronous,haarnoja2018soft}.

% RL
Many prior DRL methods tackling partially observable tasks \cite{igl2018deep,lee2019stochastic,han2019variational} are not designed for goal-driven tasks. Therefore, their learned policies are not goal-adaptive.
%where they only consider a single fixed goal that is inherently embedded in their models. 
Adapting to new goals is critical for real-world tasks, such as goal-driven visual navigation \cite{zhu2017target}, robotic object search \cite{ye2018active, yang2018visual, batra2020objectnav} and room navigation \cite{ye2018active}. Current goal-driven visual navigation methods generally neglect the essential role of estimating the agent's state under the partially observable goal-driven settings effectively, thus their performance still leaves much to be desired especially in terms of generalization ability (in-depth discussion in Section~\ref{sec:rel}). Here, we argue and show our novel GRG modeling fills the gap.

Formally, we define GRG as a complete weighted directed graph $<V, E, W>$ in which $V=G_d$ is a set of nodes representing the goals $G_d$, and $E$ is the directed edges connecting two nodes with the weights $W$. 
We incorporate GRG into HRL via two aspects: 1) weighing each candidate sub-goal in the high-level layer by $\mathbb{C}(\tau^*)$, the cost of the optimal plan $\tau^*$ from the sub-goal to the goal over the GRG; 2) terminating the low-level layer referring to the optimal plan $\tau^*$ from the proposed sub-goal to the goal over the GRG.
%and is incorporated into the HRL following ... 
To empirically validate the presented system, we start with demonstrating the effectiveness of our method in a grid-world domain where the environments are partially observable and a set of goals following a pre-defined relation are specified as the task goals. The design follows the intuition in real-world applications that certain relations hold in the goal space.  For example, in the robotic object search task, users arrange the household objects in accordance with their functionalities. Another example is the indoor navigation task where room layouts are not random. Furthermore, in addition to the grid-world experiment, we also apply our method to tackle the robotic object search task in both the AI2-THOR \cite{kolve2017ai2} and the House3D \cite{wu2018building} benchmark environments. We show HRL-GRG model exhibits superior performance in both experiments over other baseline approaches, with extensive ablation analysis.

\section{Related Work}
\label{sec:rel}
% current work and problems on partially observable goal-driven task

Research works on partially observable goal-driven tasks are explored typically under the visual navigation scenarios: an agent learns to navigate to user-specified goals with its first-person view visual inputs. Previous works' contributions lie in representation learning of the agent's underlying state and knowledge embedding for goal state inference.

In \cite{zhu2017target}, the authors present a target-driven DRL model to learn a desired action policy conditioned on both visual inputs and target goals. With a target goal being specified as an image taken at the goal position, their model captures the spatial configuration between the agent's current position and the goal position as the agent's underlying state. However, when the goal position is far away, the inputs of the model lack the information to infer the spatial configurations, and so the model instead memorizes such spatial configurations. As a result, their model relies on a scene-specific layer for every single environment. Similar issues also exist in \cite{wu2019exploring}. \cite{savinov2018semi} represents the agent's current state with respect to the goal state through a semi-parametric topological memory while it requires a pre-exploration stage to build a landmark graph. The authors of \cite{gupta2017cognitive} locate the goal position in their predicted top-down egocentric free space map. However, the method struggles when the goal is not visible. With more detailed information about the goals, the Vision-and-Language Navigation task has drawn research attention in which a fine-grained language-based visuomotor instruction serves as the goal description for the agent to follow and achieve \cite{anderson2018vision, fried2018speaker, wang2019reinforced}. Yet, specifying a goal with an image or a visuomotor instruction is inefficient and impractical for real-world applications. Instead, taking a concise semantic concept as a goal description is more desirable \cite{mousavian2019visual, yang2018visual, nguyen2019reinforcement, qiu2020target, batra2020objectnav, chaplot2020object, ye2020efficient}. 
% The semantic goals as one input to the models are typically in the forms of one-hot vectors or word embeddings. 
Semantic goals as model inputs, typically come in the form of one-hot encoded vectors or word embeddings.
Therefore, goal inputs provide limited information for estimating the agent's states relative to the goal states.

Since it is non-trivial to incorporate complete information with a goal description as input, others embed task-specific prior knowledge to infer the goal states. In \cite{yang2018visual, nguyen2019reinforcement, qiu2020target}, the authors extract object relations from the Visual Genome \cite{krishna2017visual} corpus and incorporate this prior into their models through Graph Convolutional Networks \cite{kipf2016semi}. The extracted object relations encode the co-occurrence of objects based on human annotations from the Visual Genome dataset, which may not be consistent with the target application environments and the agent's understanding of the world. More recently, the authors of \cite{wu2019bayesian} come up with the Bayesian Relational Memory (BRM) architecture to capture the room layouts of the training environments from the agent's own experience for room navigation. The BRM further serves as a planner to propose a sub-goal to the locomotion policy network. Since the proposed sub-goal is still not observable, the authors of BRM train individual locomotion polices for each sub-goal to respectively tackle one partially observable task. In such manner, the BRM model's low-level network  is not goal-adaptive and still brings about inefficiency and scaling concerns. 

%Our two-layer hierarchical model, instead,  solves a partially observable goal-driven task by decomposing it into 1) a sub-goal selection task tackled by our high-level layer with a self-learned goals relational graph, and 2) a fully observable goal-driven task handled by our low-level layer. Thus, our method is able to infer the goal state from the agent's own experience (more efficient?) and break down the original task into a set of tractable sub-tasks (goal-adaptive and scales well?).

Apart from prior research efforts, we present a novel hierarchical reinforcement learning approach equipped with a GRG formulation for the general partially observable goal-driven task. Our GRG captures the underlying relations among all goals in the goal space and enables our hierarchical model to achieve superior generalization performance by decomposing the task into a high-level sub-goal selection task and low-level fully observable goal-driven task. 
% Our GRG captures the underlying relations among all goals in the goal space through a Dirichlet-categorical model and thus enables graph-based planning. The planning outputs are further incorporated into our two-layer hierarchical RL for proposing sub-goals and early low-level layer termination.
% We validate our approach on both the grid-world domain and the challenging robotic object search task. The results show our approach is effective and is exceptional in generalizing unseen environments and new goals. We argue that the joint learning of GRG and HRL boosts the overall performance on the tasks we perform in our experiments, and it may push forward future research ventures in combining symbolic reasoning with deep RL. 

\section{Hierarchical RL with GRG}
\label{sec:method}

\subsection{Overview}
% Our focus is the partially observable goal-driven task where the agent needs to make a decision of which action to take to achieve a user-specified goal relying on its partial observations. To better illustrate our method, we take a grid-world task as an example. The agent is asked to move to a goal position between obstacles while the agent can only observe a local map of obstacles and goals. Without loss of generality, we represent the observations as images and describe the goals with categorical labels. Our objective is to learn an optimal policy for several goals and instances of this domain that can generalize to new/unseen ones.
Our focus is the partially observable goal-driven task where the agent needs to make a decision of which action to take to achieve a user-specified goal relying on its partial observations. Without loss of generality, we represent the observations $\Omega$ as images, such as the local egocentric top-down maps in the grid-world domain and the first-person view RGB images in the robotic object search task (see Figure~\ref{fig:overview} left). We specify the goal descriptions $G_d$ as categorical labels (goal indices in the grid-world domain and object categories in the robotic object search task). To better illustrate our method, we take the grid-world domain as an example. The agent is asked to move to a goal position indicated by the goal index between obstacles. The agent can only observe a local map of obstacles and goals.  The objective is to learn an optimal policy for several goals and instances of the grid-world domain which can generalize to new/unseen ones.

Figure~\ref{fig:overview} (right) depicts an overview of our method that is composed of a Goals Relational Graph (GRG), a high-level network and a low-level network. At time step $t$, the agent receives an observation $o_t \in \Omega$ that is a local map of its surrounding obstacles and a set of visible goals $VG = \{vg\ |\ vg \in G_d\ and\ \Phi(o_t, vg) > 0\}$. We take these visible goals as the candidate sub-goals at the time step $t$, and our high-level network learns a policy to select one from them to achieve the designated final goal $g \in G_d$. In order to be goal-adaptive, the system weighs each candidate sub-goal in $VG$ by its relation to the designated final goal $g$, estimated from GRG. As a result, our high-level network proposes a sub-goal $sg_t \in VG$ conditioning on both the observation $o_t$ and the designated final goal $g$. After the sub-goal $sg_t$ is proposed, our low-level network decides an action $a_t$ conditioning on both the observation $o_t$ and the sub-goal $sg_t$ for the agent to perform. Afterwards, the agent receives a new observation $o_{t+1}$, and our low-level network repeats $N_t$ times to achieve the sub-goal $sg_t$ until 1) the sub-goal $sg_t$ is achieved; 2) the low-level network terminates itself if a better sub-goal appears in its current observation; 3) the low-level network runs out of a pre-defined maximum number of steps $N_{max}^l$ . Either way, the low-level network collects an $N_t$-step long trajectory and terminates at the observation $o_{t+N_t}$. The trajectory updates the GRG. Then, the high-level network takes the control back to propose the next sub-goal. Overall, the process repeats until it either achieves the designated final goal $g$  or reaches a predefined maximum number of actions $N_{max}$.

\subsection{Goals Relational Graph (GRG)}
\label{sec:graph}
{\bf GRG representation.} We formulate GRG as a complete weighted directed graph $<V,E,W>$ on all goals in the goal space $G_d$ (i.e. $V=G_d$). For any goal $g_i$ and goal $g_j$, we define the weight $w_{ij}$ on the directed edge $(g_i, g_j)$ as a measure of how likely and quickly the goal $g_j$ would appear according to $\Phi$ if our low-level network tries to achieve the goal $g_i$. We set the weight $w_{ii}=1$ and adopt a Dirichlet-categorical model to learn $w_{ij}$ for any $i \not = j$.
 
To be specific, we first assign a random variable $X_{ij}$ to denote what would happen to the goal $g_j$ if our low-level network achieves the goal $g_i$. Every time when the goal $g_i$ is proposed by our high-level network, our low-level network generates a trajectory that has at most $N_{max}^l$ steps to achieve the goal $g_i$. It introduces the following $N_{max}^l+1$ events that $X_{ij}$ may take: 
\begin{itemize}[noitemsep]
    \item  Event $k \ (1 \le k \le N_{max}^l) $: the goal $g_j$ appears when $k$ steps are taken by our low-level network. We quantify the event $k$ as $x_{ij,k} = \gamma^{k-1}$ where $\gamma \in (0, 1]$ is the discount factor to denote how close the goal $g_j$ to the goal $g_i$.
    \item  Event ${N_{max}^l+1}$: the goal $g_j$ doesn't appear. We quantify this event as $x_{ij, N_{max}^l+1} = 0$.
\end{itemize}
It is fair to assume that $X_{ij} \sim Cat(\bm{\theta_{ij}})$ in which the parameter $\bm{\theta_{ij}} = (\theta_{ij,1}, \theta_{ij,2},...,\theta_{ij,N_{max}^l+1}) \sim Dir(\bm{\alpha_{ij}})$ is a learnable Dirichlet prior. $\bm{\alpha_{ij}} = (\alpha_{ij,1}, \alpha_{ij,2},...,\alpha_{ij, N_{max}^l+1})$ is a concentration hyperparameter representing the pseudo-counts of all event occurrences. Thus, it can be empirically chosen. Lastly,
the weight $w_{ij}$ is set as $\mathbb{E}[X_{ij}]$.

{\bf GRG update.} Each time when the low-level network is invoked to achieve the goal $g_i$, we get a trajectory as a sample $\mathcal{D}$ to update the GRG. For any goal $g_j$ in our goal space, we count the number of the occurrences of all events and denote it as $\bm{c_{ij}} = (c_{ij,1}, c_{ij,2},...,c_{ij, N_{max}^l+1})$. Since the Dirichlet distribution is the conjugate prior distribution of the categorical distribution, the posterior distribution of the parameter $\bm{\theta_{ij}}$, namely $\bm{\theta_{ij}}|\mathcal{D} \sim Dir(\bm{\alpha_{ij}}+\bm{c_{ij}})=Dir(\alpha_{ij,1}+c_{ij,1},\alpha_{ij,2}+c_{ij,2},...,\alpha_{ij,N_{max}^l+1}+c_{ij,N_{max}^l+1})$. As a result, the posterior prediction distribution of a new observation $P(X_{ij}=x_{ij,k}| \mathcal{D})$ can be estimated by Equation~\ref{eq:post_pred}, and the weight $w_{ij} = E(X_{ij}|\mathcal{D}) = \sum_kx_{ij,k}P(X_{ij}=x_{ij,k}|\mathcal{D})$.
\begin{equation}
\label{eq:post_pred}
\small
P(X_{ij}=x_{ij,k}| \mathcal{D}) = \mathbb{E}[\theta_{ij,k}|\mathcal{D}]=\frac{\alpha_{ij,k}+c_{ij,k}}{\sum_k(\alpha_{ij,k}+c_{ij,k})}.
\end{equation}

{\bf GRG planning.} With the GRG being learned and updated, we quantify the relation of a goal $g_i$ to a goal $g_j$ by the cost $\mathbb{C}(\tau_{i,j}^*)$ of the optimal plan $\tau_{i,j}^*$ searched from $g_i$ to $g_j$ over the GRG. In particular, suppose $\tau_{i,j} = \{\tau_1, \tau_2,...,\tau_M\}$ is a plan searched from $g_i$ to $g_j$ over the GRG in which $g_{\tau_m \ (1 \le m \le M)}$ is a goal from our goal space $G$, $\tau_1 = i$ and $\tau_M = j$, we define the optimal plan $\tau_{i,j}^* = \argmax_{\tau_{i,j}} \prod_{m=1}^{M-1}w_{\tau_m \tau_{m+1}}$. We adopt the cost of the optimal plan $\tau_{i,j}^*$, $\mathbb{C}(\tau_{i,j}^*) = \max_{\tau_{i,j}} \prod_{m=1}^{M-1}w_{\tau_m \tau_{m+1}} $ as the measure of the relation from the goal $g_i$ to the goal $g_j$.

\subsection{Goal-driven High-level Network}
% reward + Q-learning
{\bf Model formulation.} The high-level network selects a sub-goal $sg$ aiming to achieve the designated final goal $g$. We first introduce an extrinsic reward $r^e$. Here, we adopt a binary reward as the extrinsic reward to encourage the agent to achieve the final goal. Specifically, the agent receives a reward of $1$ if it achieves the final goal $g$, i.e. $r_t^e(s_{t-1}, a_{t-1}, s_t, g) = 1$ iff the state $s_t$ is the goal $g$'s state, and $0$ otherwise. Thus, the high-level task is formulated as maximizing the Q-value $Q_h^e(s,g,sg) =  \mathbb{E}[\sum_{t}^{\infty}\gamma^t r^e_{t+1} | s_t=s,g=g,sg_t=sg]$, which is the discounted cumulative extrinsic rewards expected over all trajectories starting at the state $s_t$ and the sub-goal $sg_t$. To approximate $Q_h^e(s,g,sg)$, we adopt the Q-learning technique \cite{mnih2015human} to update the parameters of the high-level network $\theta_h$ by Equation~\ref{eq:h_update}, where $R_1^e = r^e(s,a,s',g)+\gamma \max_{sg'}Q_h^e(s',g, sg')$ is the $1$-step extrinsic return. The sub-goal $sg$ is given by $\argmax_{sg}Q_h^e(s,g,sg)$ towards achieving the final goal $g$.
\begin{equation}
\label{eq:h_update}
\small
\theta_{h} \gets \theta_{h} - \nabla_{\theta_{h}}(R_1^e-Q_{\theta_{h}}(s,g,sg))^2.
\end{equation}

{\bf Network architecture.} To approximate $Q_h^e(s,g,sg)$, we condition the high-level network on the state $s$, goal $g$ and the sub-goal $sg$. A widely adopted way is by taking the state $s$ and the goal $g$ as the inputs, and output Q-values that each of them corresponds to a candidate sub-goal $sg$. Here, since the state $s$ is unknown, we instead take the observation $o$ as the input to our high-level network attempting to estimate the state $s$ simultaneously. To ensure the sub-goal that can be achieved by the low-level network, the sub-goal space at the time $t$ is set as the observable  goals within the observation $o_t$. \footnote{In practice, we supplement a back-up ``\textit{random}'' sub-goal driving the low-level network to randomly pick an action to perform in case no observable goals available. } As a consequence, the sub-goal space varies at each time stamp and is typically much smaller than the goal space. Thus, it is not efficient for the high-level network to calculate as many Q-values as the size of the goal space. Instead, as the sub-goal space is self-contained in the observation $o_t$, we hereby extract the information of each candidate sub-goal $sg$ from the observation $o_t$ and feed it into the high-level network to output one single Q-value $Q_h^e(s,g,sg)$ specifically for the sub-goal $sg$. 

Last but not the least, although most prior methods directly take the goal description $g_d$ as an additional input to their networks, we notice that the goal description $g_d$, typically in the form of a one-hot vector or word embedding, does not directly provide any information for either inferring the goal state or determining a quality sub-goal. Therefore, we opt to correlate the goal $g$ with each candidate sub-goal $sg$ by their relations. Here, our system plans over the GRG and gets the cost $\mathbb{C}(\tau_{sg,g}^*)$ of the optimal plan $\tau_{sg,g}^*$ from the sub-goal $sg$ to the goal $g$ as described in Section~\ref{sec:graph}. We multiply the cost $\mathbb{C}(\tau_{sg,g}^*)$ to the sub-goal input $sg$ elementwise before feeding it into the high-level network to predict its Q-value $Q_h^e(s,g,sg)$. In such a way, $Q_h^e(s,g,sg) = Q_h^e(s, sg\odot\mathbb{C}(\tau_{sg,g}^*))$ where the goal $g$ is embedded with beneficial information for the Q-value prediction and the sub-goal selection. As the inputs and the outputs are specified, the remaining architecture of our high-level network is flexible per application.

\begin{figure*}[ht]
\centering
\begin{tabular}{ccc}
\subfloat[optimal trajectory]{\includegraphics[width=0.28\textwidth]{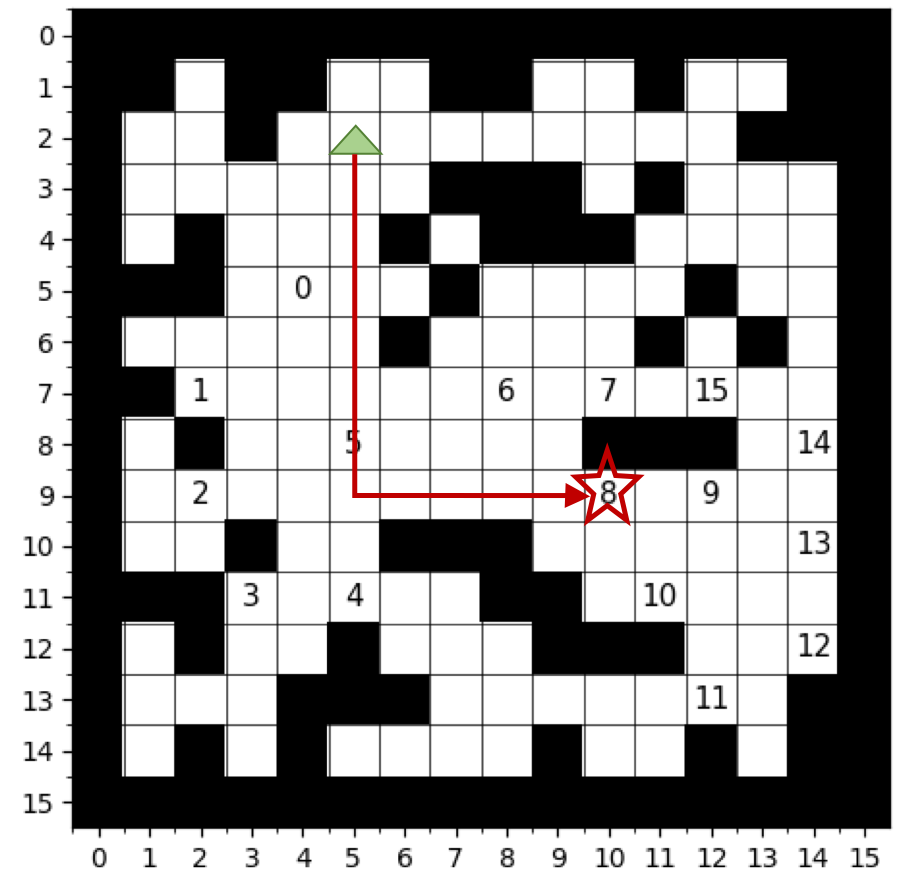}}
&\subfloat[trajectory w.o. termination]{\includegraphics[width=0.28\textwidth]{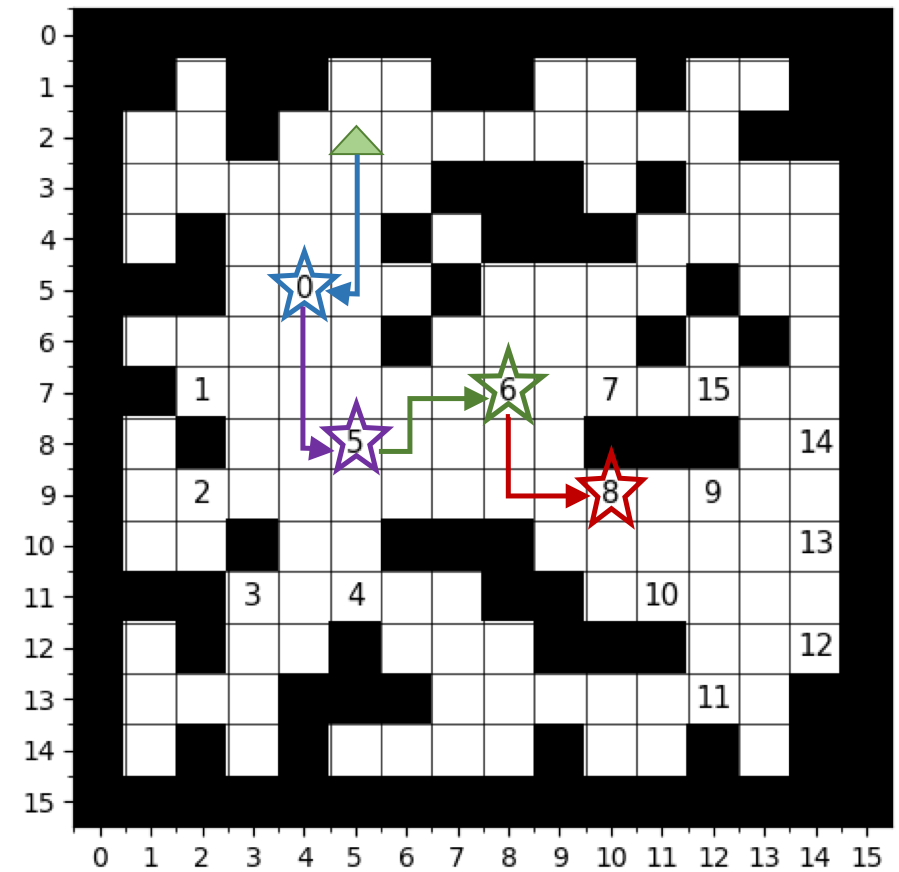}}
&\subfloat[trajectory with termination]{\includegraphics[width=0.28\textwidth]{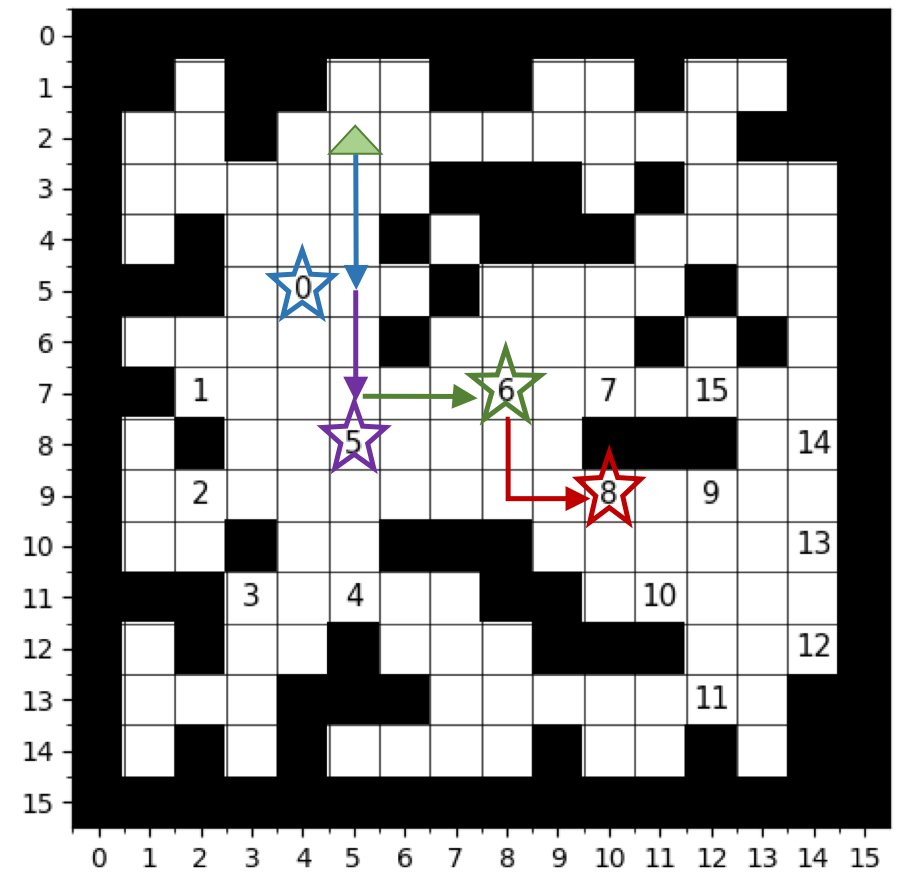}}
\end{tabular}
\caption{An illustration of how termination helps. The green triangle denotes the starting position. The stars and the arrows with different colors represent different sub-goals and the corresponding sub-goal-oriented trajectories. Termination helps to express an optimal trajectory with the limited sub-goal space.}
\label{fig:termination}
\end{figure*}

\subsection{Termination-aware Low-level Network}
% reward + Q-learning
{\bf Model formulation.} The objective of the low-level network is to learn an optimal action policy to achieve the proposed sub-goal $sg$. Similar to the high-level network, we adopt a binary intrinsic reward $r^i$ accordingly that $r_t^i(s_{t-1},a_{t-1},s_t,sg) = 1$ iff our low-level network achieves the sub-goal $sg$, and is otherwise $0$. The optimal action policy can then be learned by maximizing the expected discounted cumulative intrinsic rewards $\mathbb{E}[\sum_t^\infty \gamma^t r_{t+1}^i|s_t, sg_t, a_t]$. Since the proposed sub-goal $sg_t$ is guaranteed to be observable in the observation $o_t$, we have a fully observable goal-driven task that can be efficiently solved by the state-of-the-art reinforcement learning algorithms \cite{mnih2015human,mnih2016asynchronous,haarnoja2018soft}.

Adopting a hierarchical model to decompose a complex task into a set of sub-goal-driven simple tasks has been proven to be efficient and effective \cite{levy2017learning, nachum2018data}. Still, it is under the assumption that the goal/sub-goal space is identical to the state space so that any optimal trajectory can be expressed by a sequence of optimal sub-goal-oriented trajectories. In our work, we consider a practical setting in which the goal/sub-goal space is much smaller than the state space, as lots of intermediate states are not of interest in terms of solving the task. Consequently, an optimal trajectory may not be expressed by the limited presented sub-goals on the trajectory as Figure~\ref{fig:termination} (a) shows. Instead, following the set of available sub-goals proposed could yield a less optimal trajectory as Figure~\ref{fig:termination} (b) depicts.

% \begin{wrapfigure}{r}{.4\textwidth}
% \setlength{\unitlength}{0.4\textwidth}
% \centering
% \vspace{-1.5em}
% \includegraphics[width=0.4\textwidth]{figures/relation_graph.png}
% \caption{A visualization of a GRG learned on the grid-world domain ($g_{16}$ is the back-up ``\textit{random}'' goal).}
% \vspace{-2em}
% \label{fig:relation}
% \end{wrapfigure}

\begin{figure}[ht!]
\centering
\includegraphics[width=0.38\textwidth]{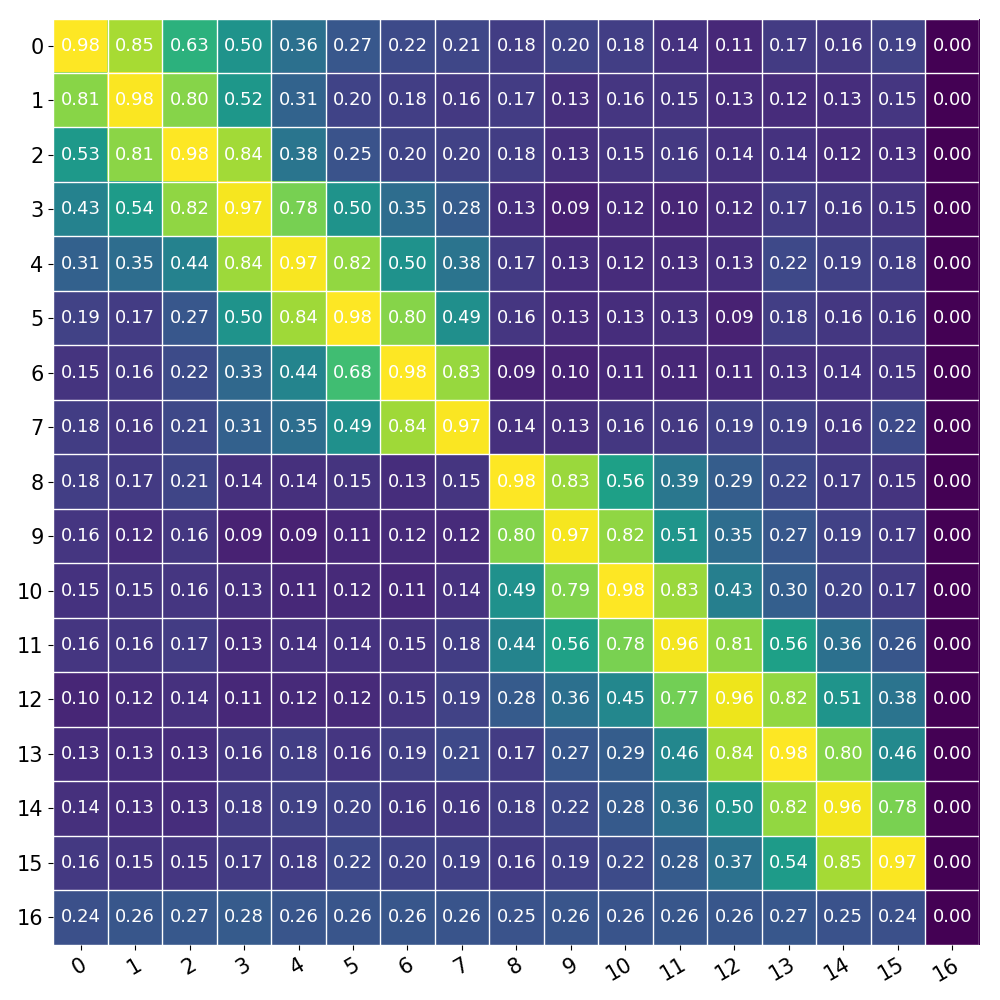}
\caption{A visualization of a GRG learned on the grid-world domain ($g_{16}$ is the back-up ``\textit{random}'' goal).}
\vspace{-8pt}
\label{fig:relation}
\end{figure}

To overcome this issue, we further allow the low-level network to terminate at a valuable state before it achieves the proposed sub-goal. The intuition is that along its way to the sub-goal, the agent may reach a state that is better poised for achieving the final goal. Namely, a state where a better sub-goal appears (see Figure~\ref{fig:termination} (c)). Some prior methods explore modeling a termination function in their formulations \cite{bacon2017option} or adding a special ``stop'' action in the action space for an optimal stop policy \cite{yang2018visual}. However, they unavoidably increase the exploration difficulty and hurt the sample efficiency. Instead, we terminate our low-level network under the supervision of GRG. Whenever a sub-goal $sg$ is received, an optimal plan $\tau_{sg,g}^*$ starting from the sub-goal $sg$ to the goal $g$ over the GRG is generated following Section~\ref{sec:graph}. In fact, any goal on the $\tau_{sg,g}^*$ other than $sg$ is a better sub-goal for achieving the final goal $g$, and once it appears, our low-level network terminates and returns the control back to the high-level network.

\begin{table*}
\small
\caption{The performance of all methods on the unseen gird-world maps.}
\label{tbl:baseline}
\begin{center}
\begin{tabular}{lccccccccccc}
\specialrule{0.12em}{0pt}{2pt}
& \multicolumn{3}{c}{Seen Goals}&& \multicolumn{3}{c}{Unseen Goals}&& \multicolumn{3}{c}{Overall}\\
\cline{2-4}\cline{6-8}\cline{10-12}
\specialrule{0em}{1pt}{1pt}
Method &SR$\uparrow$ &AS / MS$\downarrow$ & SPL$\uparrow$ &&SR$\uparrow$ &AS / MS$\downarrow$ & SPL$\uparrow$&&SR$\uparrow$ &AS / MS$\downarrow$  & SPL$\uparrow$ \\
\specialrule{0.12em}{1pt}{1.5pt}
\textsc{Oracle} & 1.00 &11.81 / 11.81 &1.00 &&1.00 &11.28 / 11.28 & 1.00 && 1.00 &10.38 / 10.38 &1.00  \\
\specialrule{0.08em}{1pt}{1pt}
\textsc{Random} &0.16 &42.15 / 5.47 &0.03 &&0.15 &42.38 / 4.81 &0.04 &&0.18 &36.62 / 4.69 &0.05 \\
\textsc{DQN} &0.20 &20.28 / 5.47 &0.13 &&0.20 &11.90 / 4.10 &0.15 &&0.32 &16.23 / 5.71 &0.23 \\
\textsc{h-DQN} &0.43 &\textbf{20.25 / 7.95} & 0.28 &&0.19 &26.09 / 6.38 &0.08 &&0.45 &20.84 / 7.16 &0.26 \\
\specialrule{0.01em}{1pt}{1pt}
\textbf{Ours} &\textbf{0.57} &28.71 / 9.03 &\textbf{0.33} &&\textbf{0.70} &\textbf{24.19 / 8.73} &\textbf{0.45} &&\textbf{0.74} &\textbf{24.02 / 8.65} &\textbf{0.46} \\
\specialrule{0.12em}{1.5pt}{0pt}
\end{tabular}
\end{center}
\end{table*}

{\bf Network architecture.} We implement the termination mechanism in the low-level policy using the GRG which is decoupled from low-level policy learning. Therefore, the low-level network still addresses the standard fully observable goal-driven task, i.e. predicting the optimal action policy from the current observation $o_t$ that includes the information of the sub-goal $sg_t$. This can be solved by methods like DQN \cite{mnih2015human} and A3C \cite{mnih2016asynchronous}, without special requirements on the network architecture.

\section{Experiments}

Our experiments aim to seek the answers to the following research questions, 1) Is GRG able to capture the underlying relations of all goals? 2) Is GRG able to help solve the new, unseen partially observable goal-driven tasks, and if yes, how? 3) How well does the proposed method work for the goal-driven visual navigation task? To answer the first two questions, we conduct evaluation in an unbiased synthetic grid-world domain. To answer the third question, we apply our system on both AI2-THOR \cite{kolve2017ai2} and House3D \cite{wu2018building} environments for the robotic object search task . \footnote{For technical implementation details, please refer to the supplementary material.}

\subsection{Grid-world Domain}
\label{sec:grid-world}
{\bf Grid-world generation.} We generate a total of $120$ grid-world maps of size $16 \times 16$ with randomly placed obstacles taking up around $35\%$ of the space. We arrange $16$ goals in the free spaces of each map following a pre-defined pattern to test if our proposed GRG can capture it. Specifically, we randomly place goal $g_0$ and goal $g_8$ first. Then, for $0<i<16$ and $i\not=8$, we place goal $g_i$ at a random place in the window of size $7\times7$ centered at goal $g_{i-1}$. Figure~\ref{fig:overview} shows an instance of a grid-world map. We take $100$ grid-world maps and $12$ goals for training, with the remaining $20$ grid-world maps and the corresponding $16$ goals are kept for testing.

{\bf Baseline methods.} We assume the agent can only observe the window of size $7\times 7$ centered at its position, which is represented by an image including the map of obstacles and any goal positions. The agent can take one action as moving up/down/left/right, and would stay at the current position if the action leads to collision. Success is defined as the agent reaches the position of the designated goal. For this task, we adopt the DQN \cite{mnih2015human} algorithm for our low-level network to learn the optimal action policy to achieve the sub-goal proposed by our high-level network, and we compare our method with the following baseline methods.
\begin{itemize}[noitemsep]
    \item \textsc{Oracle} and \textsc{Random}. The agent always takes the optimal action or a random action respectively. The two methods are taken as performance upper/lower bounds.
    \item \textsc{DQN}. The vanilla DQN implementation that directly maps the observation to the optimal action. To make it goal-adaptive, the input observation image contains a channel of the obstacle map and a channel of the designated goal position if it presents. It is empirically shown to be better than embedding the goal with a one-hot vector (see supplementary material).
    \item \textsc{h-DQN}. It is a widely adopted hierarchical method \cite{kulkarni2016hierarchical} modified for our partially observable goal-driven task where both the high-level network and the low-level network adopt a vanilla DQN implementation. The high-level network takes the whole observation as the input to propose a sub-goal that is visible. To be goal-adaptive, the goal is embedded into the high-level network in the form of a one-hot encoded vector. The low-level network is the same as the method \textsc{DQN} (also with ours).
\end{itemize}

\begin{table*}
\small
\caption{The ablation studies of our method on the unseen gird-world maps.}
\label{tbl:ablation}
\begin{center}
\begin{tabular}{lccccccccccc}
\specialrule{0.12em}{0pt}{2pt}
& \multicolumn{3}{c}{Seen Goals}&& \multicolumn{3}{c}{Unseen Goals}&& \multicolumn{3}{c}{Overall}\\
\cline{2-4}\cline{6-8}\cline{10-12}
\specialrule{0em}{1pt}{1pt}
Method &SR$\uparrow$ &AS / MS$\downarrow$ & SPL$\uparrow$ &&SR$\uparrow$ &AS / MS$\downarrow$ & SPL$\uparrow$&&SR$\uparrow$ &AS / MS$\downarrow$  & SPL$\uparrow$ \\
\specialrule{0.12em}{1pt}{1.5pt}
\textbf{Ours} &\textbf{0.57} &\textbf{28.71 / 9.03} &\textbf{0.33} &&\textbf{0.70} &24.19 / 8.73 &\textbf{0.45} &&\textbf{0.74} &\textbf{24.02 / 8.65} &\textbf{0.46} \\
\specialrule{0.08em}{1pt}{1pt}
 -relation &0.26 &33.20 / 6.16 &0.10 &&0.35 &31.93 / 6.84 &0.14 &&0.40 &29.39 / 6.14 &0.18 \\
 -termination &0.55 &31.36 / 8.81 & 0.27 &&0.58 &27.91 / 8.11 &0.32 &&0.64 &25.56 / 7.88 &0.37 \\
 -high-level  &0.56 &29.97 / 9.03 &0.31 &&0.65 &\textbf{23.63 / 8.65} &0.42 &&0.66 &22.86 / 7.86 &0.41 \\
\specialrule{0.12em}{1.5pt}{0pt}
\end{tabular}
\end{center}
\end{table*}

{\bf Baseline comparisons.} We specify the maximum number of actions that all methods can take as $100$, and for hierarchical methods, i.e. \textsc{h-DQN} and our method, the maximum number of actions that the low-level network can take at each time is $10$. We evaluate all methods in terms of the Success Rate (SR), the Average Steps over all successful cases compared to the Minimal Steps over these cases (AS / MS), and the Success weighted by inverse Path Length (SPL) following \cite{anderson2018evaluation} and calculated as $\frac{1}{N}\sum_{i=1}^{N}S_i \frac{l_i}{\max(l_i, p_i)}$. Here $S_i$ is a binary indicator of success in experiment $i$, $l_i$ and $p_i$ are the minimal steps and the steps actually taken by the agent. We randomly sample seen goals, unseen goals and all goals over the unseen grid-world maps, each having $100$ samples that yield $100$ tasks respectively. We run each method using $5$ random seeds. Table~\ref{tbl:baseline} reports the results.

As is shown in Table~\ref{tbl:baseline}, we can observe that our method outperforms all baseline methods in terms of generalization ability on the unseen grid-world maps as expected. On one hand, the performance of \textsc{DQN} leaves much to be desired for both seen goals and unseen goals. Whereas \textsc{h-DQN} achieves comparable performance to our method for seen goals, but it struggles to generalize towards unseen goals. On the other hand, our proposed method generalizes well to both seen goals and unseen goals, since our GRG captures the underlying relations of all goals, even if some of the goals are not set as the designated goals in the training stage. Figure~\ref{fig:relation} shows a visualization of the learned GRG, which captures the goal relations well.

{\bf Ablation studies.} To investigate \textit{how} GRG helps to solve the partially observable goal-driven task, we conduct ablation studies for each component. The GRG has two roles: In the high-level network, it weighs each candidate sub-goal by its relation to the final goal before calculating its Q-value. In the low-level network, it is used for early termination. We disable each role and denote them as ``-relation'' and ``-termination'' respectively. The results reported in Table~\ref{tbl:ablation} clearly show that both of them contribute to the performance of our proposed method, whereas weighing the candidate sub-goals by relations contributes more. Moreover, to show the necessity of the high-level network, we present ``-high-level'' that removes the high-level network, leaving only the GRG and the low-level network in place. In such a way, a sub-goal is proposed purely based on the graph planning over GRG without taking the current observation into consideration. The results in Table~\ref{tbl:ablation} show that it is slightly worse than our proposed method; from which we can infer that 1) the high-level network captures as much information as the GRG; 2) observations still matter since the graph only captures the expected relations; and 3) the performance gap could be wider in complex real-world environments.

\subsection{Robotic Object Search}

\begin{table*}
\small
\caption{The performance of all methods in the House3D \cite{wu2018building} environment for the robotic object search task.}
\label{tbl:ros}
\begin{center}
\begin{tabular}{lccccccccccc}
% \specialrule{0.12em}{0pt}{2pt}
& \multicolumn{5}{c}{Single Environment} && \multicolumn{5}{c}{Multiple Environments} \\
\cline{2-6}\cline{8-12}
& \multicolumn{2}{c}{Seen Goals}&& \multicolumn{2}{c}{Unseen Goals}&& \multicolumn{2}{c}{Seen Env.}&& \multicolumn{2}{c}{Unseen Env.}\\
\cline{2-3}\cline{5-6}\cline{8-9}\cline{11-12}
\specialrule{0em}{1pt}{1pt}
Method &SR$\uparrow$ &SPL$\uparrow$ & &SR$\uparrow$ & SPL$\uparrow$& &SR$\uparrow$ & SPL$\uparrow$ &&SR$\uparrow$ & SPL$\uparrow$ \\
\specialrule{0.12em}{1pt}{1.5pt}
\textsc{Random} &0.20 &0.05 &&0.23 &0.04 &&0.39 &0.03 &&0.60 &0.05 \\
\textsc{DQN} &0.58 &0.27 &&0.18 &0.05 &&0.42 &0.06 &&0.39 &0.04 \\
\textsc{A3C} &0.53 &0.18 &&0.27 &0.09 &&0.48 &0.03 &&0.47 &0.03 \\
\textsc{Hrl} &0.77 &0.15 &&0.05 &0.00 &&0.43 &0.05 &&0.28 &0.02 \\
\specialrule{0.01em}{1pt}{1pt}
\textbf{Ours} &\textbf{0.88} &\textbf{0.33} &&\textbf{0.79} &\textbf{0.21} &&\textbf{0.76}  &\textbf{0.20}&&\textbf{0.62} &\textbf{0.10} \\
\specialrule{0.12em}{1.5pt}{0pt}
\end{tabular}
\end{center}
\end{table*}

%Among many real-world applications where the goals are related by nature, we show how our proposed method works for the 
Robotic object search is a challenging goal-driven visual navigation task \cite{yang2018visual, ye2018active, mousavian2019visual, nguyen2019reinforcement, ye2019gaple, qiu2020target,batra2020objectnav}. It requires an agent to search for and navigate to an instance of a user-specified object category in indoor environments with only its first-person view RGB image. 
%We conduct experiments on the simulation platform House3D \cite{wu2018building}. 

A previous method \textsc{Scene Priors} \cite{yang2018visual} also incorporates object relations as scene priors to improve the robotic object search performance in the AI2-THOR \cite{kolve2017ai2} environments. Unlike ours, it extracts the object relations from the Visual Genome \cite{krishna2017visual} corpus and incorporates the relations through Graph Convolutional Networks \cite{kipf2016semi}. Therefore, we compare our method with it in the AI2-THOR environments. AI2-THOR consists of $120$ single functional rooms, including kitchens, living rooms, bedrooms and bathrooms, in which we take the first-person view semantic segmentation and depth map as the agent's pre-processed observation. As such, the goal position can be represented by the corresponding channel of the semantic segmentation (a.k.a. $\Phi$). In addition, we adopt the A3C \cite{mnih2016asynchronous} algorithm for our low-level network and define the maximum steps it can take at each time as $10$. We follow the experimental setting in \cite{yang2018visual} to implement both \textsc{Scene Priors} \cite{yang2018visual} and our \textsc{HRL-GRG}. We report the results in Table~\ref{tbl:ros_ai2thor} where we compare the two methods in terms of their performance improvement over the \textsc{Random} method.
% Since the test samples (i.e. rooms, target objects and the agent's initial positions) used in \cite{yang2018visual} are unknown, we hereby compare them in terms of the performance improvement over their corresponding \textsc{Random} methods. 
Table~\ref{tbl:ros_ai2thor} indicates an overfitting issue of the \textsc{Scene Priors} method as reported in \cite{yang2018visual} as well. At the same time, we observe a superior generalization ability of our method especially to the unseen goals.
% even though we have a unifying model for all room types while \textsc{Scene Priors} has separating models for each room type.

\begin{table}[h]
\small
\caption{The performance improvement of \textsc{Scene Priors} \cite{yang2018visual} (top) and our \textsc{HRL-GRG} (bottom) over the \textsc{Random} method in the AI2-THOR \cite{kolve2017ai2} environment for the robotic object search task (without stop action).}
    \centering
    \begin{tabular}{ccccccc}
    & & \multicolumn{2}{c}{Seen Goals}&& \multicolumn{2}{c}{Unseen Goals}\\
    \cline{3-4}\cline{6-7}
    & & SR$\uparrow$ &SPL$\uparrow$ & &SR$\uparrow$ & SPL$\uparrow$\\
    \specialrule{0.12em}{1pt}{1.5pt}
    \multirow{2}{*}{Seen Env.} &  \cite{yang2018visual}   &+0.25 &+0.16 && +0.08 &+0.07\\
    & \textbf{Ours} &\textbf{+0.37} &\textbf{+0.24} &&\textbf{+0.33} &\textbf{+0.23}\\
    \specialrule{0.01em}{1pt}{1pt}
    \multirow{2}{*}{Unseen Env.} &   \cite{yang2018visual} &+0.18 &+0.11 && +0.12 &+0.06 \\
   & \textbf{Ours} &\textbf{+0.33} &\textbf{+0.21} &&\textbf{+0.38} &\textbf{+0.23}\\
   \specialrule{0.12em}{1pt}{1.5pt}
         
    \end{tabular}
    \label{tbl:ros_ai2thor}
\end{table}

% House3D simulation platform \cite{wu2018building} is a widely adopted test bed. It provides a large number of house environments, each of which has multiple functional rooms with various objects being placed. They may occlude the user-specified target object, thus the agent needs to infer the target object's location on the fly.

To further demonstrate the efficacy of our method in more complex environments, we conduct robotic object search on the House3D \cite{wu2018building} platform. Different from AI2-THOR, each house environment in the House3D has multiple functional rooms that are more likely to occlude the user-specified target object, thus stressing upon the ability of inferring the target object's location on the fly to perform the task well.

We consider a total of $78$ object categories in the House3D environment to form our goal space. The agent moves forward / backward / left / right $0.2$ meters, or rotates $90$ degrees for each action step. 
% To equip the agent with sensing capabilities, 
We adopt the encoder-decoder model from \cite{chen2018encoder} to predict both the semantic segmentation and the depth map from the first-person view RGB image and we take both predictions as the agent's partial observation. 
Furthermore, we adopt the A3C \cite{mnih2016asynchronous} algorithm for the low-level network. 
% to reach the position of the proposed sub-goal, taking both the predicted depth map and the sub-goal specified channel of the predicted semantic segmentation as the inputs.
We compare our method with the baseline methods introduced in Section~\ref{sec:grid-world} while we also adopt the A3C algorithm for the low-level network in \textsc{h-DQN} and hereby denoted as \textsc{Hrl}. In addition, we include the vanilla \textsc{A3C} approach.

\begin{figure}[ht!]
\centering
\includegraphics[width=\columnwidth]{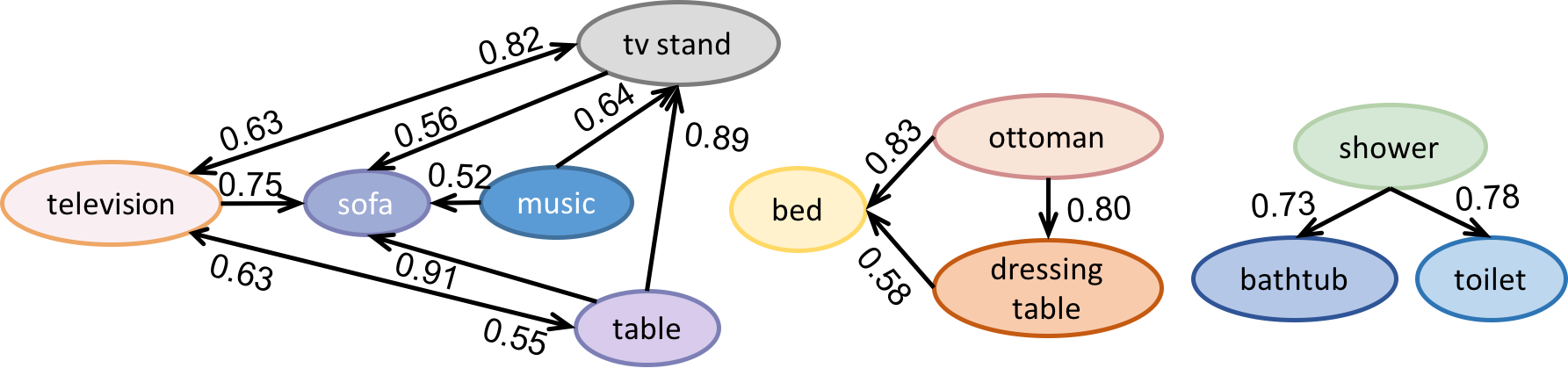}
\caption{The object relations captured by our GRG in the House3D \cite{wu2018building} environment for the robotic object search task. Only a small number of objects as nodes and the edges with the weight $\ge 0.5$ are shown.}
\label{fig:obj_relation}
\end{figure}

We set the maximum steps for all the aforementioned methods to solve the object search task in the House3D environment as $1000$, and the maximum steps that the low-level networks of the hierarchical methods (\textsc{Hrl} and ours) can take as $50$. To better investigate each method's properties, we first train and evaluate in a single environment, and show the results in Table~\ref{tbl:ros} (left part). Similar to the grid-world domain, the baseline methods lack generalization ability towards achieving the unseen goals, even though they perform fairly well for the seen ones. The placement of many objects is subject to the users' preference that may require the environment-specific training process. Still, it is desirable for a method to generalize towards the objects in the unseen environments where the placement of the objects is consistent with that in the seen ones (e.g., the objects that are always placed in accordance with their functionalities). We train all the methods in four different environments and test the methods in four other unseen environments. The results presented in Table~\ref{tbl:ros} (right part) show that all the baselines struggle with the object search task under multiple environments even during the training stage. In comparison, our method achieves far superior performance with the help of the object relations captured by our GRG (samples shown in Figure~\ref{fig:obj_relation}).

\section{Conclusion}
In this paper, we present a novel hierarchical reinforcement learning approach equipped with a GRG formulation for the partially observable goal-driven task. Our GRG captures the underlying relations among all goals in the goal space through a Dirichlet-categorical model and thus enables graph-based planning. The planning outputs are further incorporated into our two-layer hierarchical RL for proposing sub-goals and early low-level layer termination. We validate our approach on both the grid-world domain and the challenging robotic object search task. The results show our approach is effective and is exceptional in generalizing to unseen environments and new goals. We argue that the joint learning of GRG and HRL boosts the overall performance on the tasks we perform in our experiments, and it may push forward future research ventures in combining symbolic reasoning with DRL. 

\noindent{\bf Acknowledgements. }
This work is partially supported by the NSF grant \#1750082, and Samsung Research.

{
\newpage
\small
\bibliographystyle{ieee_fullname}
\bibliography{reference}
}

\newpage
\appendix
\section{Experiments in the Grid-world Domain}
\label{appendix:gw}
\subsection{Network Architectures}
\label{appendix:gw_arch}
The detailed network architectures of all the baseline methods and our method are described in the following.
\begin{itemize}
    \item \textsc{DQN}. The vanilla DQN implementation that takes the observation of shape $7\times 7 \times 2$ as the input in which the first channel represents the obstacles and the second channel denotes the designated goal position. The input is fed into three convolutional layers with $1$, $50$ and $100$ filters of kernel size $1\times 1$, $3 \times 3$ and $3\times 3$ respectively where a $2\times 2$ max-pooling is attached after the second and the third layer. It is further followed by two fully-connected layers with $100$ hidden units and $4$ outputs respectively. Each output corresponds to a Q-value $Q(s,g,a)$ of an action. 
    
    Since the goal may not be observable all the time, i.e. the second channel of the input may not have the goal information, we also explore the implementation \textsc{DQN\_onehot} where we specify the goal with the $16$ dimensional one-hot vector and we concatenate it with the input vector of the first fully-connected layer in the method \textsc{DQN}, and the implementation \textsc{DQN\_full} similar to the \textsc{DQN\_onehot} while we take the full observation of shape $7 \times 7 \times 17 $ as the input. The first channel of the input is the obstacle map and each of the remaining $16$ channels denotes corresponding goal position. We find both \textsc{DQN\_onehot} and \textsc{DQN\_full} perform worse than \textsc{DQN} as Table~\ref{tbl:dqn} shows. Therefore, we compare our method with \textsc{DQN}. 
    
    \item \textsc{h-DQN}. The hierarchical method where the high-level network takes the full observation of shape $7 \times 7 \times 17 $ as the input and consists of three convolutional layers ($1$ filter of kernel size $1\times 1$, $50$ and $100$ filters of kernel size $3 \times 3$ with $2 \times 2$ max-pooling after the second and the third layers), and two fully-connected layers with $100$ hidden units and $17$ outputs respectively. The goal that in the form of the $16$ dimensional one-hot vector is concatenated with the hidden vector before inputting to the first fully-connected layer, and the $17$ outputs are the Q-values $Q_h^e(s,g,sg)$ for all possible sub-goals including the back-up ``random'' sub-goal. The low-level network is exactly same as the method \textsc{DQN} in which the second channel of the input represents the position of the proposed sub-goal.
    
    \item \textsc{Ours}. The high-level network of our method \textsc{HRL-GRG} is similar to the method \textsc{DQN} while the second channel of the input denotes the position of a candidate sub-goal (a goal that is observable) and only a single output is generated to represent the Q-value $Q_h^e(s,g,sg)$ for the input sub-goal. Our low-level network is exactly same as the method \textsc{DQN} and the low-level network of \textsc{h-DQN}.

\end{itemize}

\begin{table*}[h]
% \small
\caption{The performance of \textsc{DQN} vs \textsc{DQN\_onehot} and \textsc{DQN\_full} on the unseen gird-word maps.}
\label{tbl:dqn}
\begin{center}
\begin{tabular}{lccccccccccc}
\specialrule{0.12em}{0pt}{2pt}
& \multicolumn{3}{c}{Seen Goals}&& \multicolumn{3}{c}{Unseen Goals}&& \multicolumn{3}{c}{Overall}\\
\cline{2-4}\cline{6-8}\cline{10-12}
\specialrule{0em}{1pt}{1pt}
Method &SR$\uparrow$ &AS / MS$\downarrow$ & SPL$\uparrow$ &&SR$\uparrow$ &AS / MS$\downarrow$ & SPL$\uparrow$&&SR$\uparrow$ &AS / MS$\downarrow$  & SPL$\uparrow$ \\
\specialrule{0.12em}{1pt}{1.5pt}
\textsc{DQN} &0.20 &20.28 / 5.47 &0.13 &&0.20 &11.90 / 4.10 &0.15 &&0.32 &16.23 / 5.71 &0.23 \\
\textsc{DQN\_onehot} &0.03 &50.85 / 6.47 & 0.00 &&0.05 &26.86 / 4.10 &0.02 &&0.03 &36.66 / 3.13 &0.01 \\
\textsc{DQN\_full} &0.01 &24.85 / 3.05 & 0.00 &&0.05 &31.53 / 5.75 &0.03 &&0.05 &23.55 / 3.15 &0.03 \\

\specialrule{0.12em}{1.5pt}{0pt}
\end{tabular}
\end{center}
\end{table*}

\subsection{Training Protocols and Hyperparameters}
\label{appendix:gw_train}
 For all the networks, we adopt the Double DQN \cite{van2016deep} technique and we train all the methods on the $100$ training grid-word maps to achieve the $12$ goals ($g_0$, $g_1$, $g_3$, $g_4$, $g_6$, $g_7$, $g_8$, $g_9$, $g_{11}$, $g_{12}$, $g_{14}$ $g_{15}$). We first train the method \textsc{DQN} to achieve the goal from where the goal is observable and we take the model as the pre-trained model for the \textsc{DQN} and the low-level networks of both \textsc{h-DQN} and our method. For all the methods, we adopt the curriculum training paradigm. To be specific, at episode $i<10000$, we start the agent at a position that is randomly selected from the top $(i+10)/100\ \%$ positions closest to the goal position, and when $i \ge 10000$, we start the agent at a random position. All the hyperparameters are summarized in Table~\ref{tbl:gw_hp}. We evaluate all the methods on the $20$ testing grid-world maps over $5$ different random seeds, namely $1$, $5$, $13$, $45$ and $99$.
\begin{table*}
    % \small
    \caption{Hyperparameters of all the methods for the grid-world domain.}
    \label{tbl:gw_hp}
    \centering
    \begin{tabular}{lll}
        \specialrule{0.12em}{0pt}{2pt}
        Hyperparameter & Description & Value\\
         \specialrule{0.05em}{0pt}{2pt}
         $\gamma$ & Discount factor & 0.99\\
         lr & Learning rate for all networks (high-level/low-level) & 0.0001\\
         main\_update & Interval of updating all the main networks of Double DQN & 10 \\
         target\_update & Interval of updating all the target networks of Double DQN & 10000\\
         batch\_size & Batch size for training all the networks & 64 \\
         epsilon & Initial exploration rate, anneal episodes, final exploration rate & 1, 10000, 0.1 \\ 
         max\_episodes & Maximum episodes to train each method & 100000\\
         $N_{max}^l$ & The maximum steps that low-level network can take if applied & 10 \\
         $N_{max}$ & The maximum steps that each method can take & 100 \\
         optimizer & Optimizer for all the networks & RMSProp \\
         $\bm{\alpha_{ij}}$ & The hyperparameter of our GRG $(\alpha_{ij,1},...,\alpha_{ij,10},\alpha_{i,j,11})$ & $(0,...,0,1)$\\
         \specialrule{0.12em}{0pt}{2pt}
    \end{tabular}
\end{table*}

\subsection{Qualitative Results}
We show some qualitative results performed by our method on the unseen grid-world maps to achieve both seen goals and unseen goals in Figure~\ref{fig:gw_results}. The results shall be better viewed in the supplementary videos.
\begin{figure*}[ht]
\centering
\begin{tabular}{cc}
\subfloat[]{\includegraphics[width=0.47\textwidth]{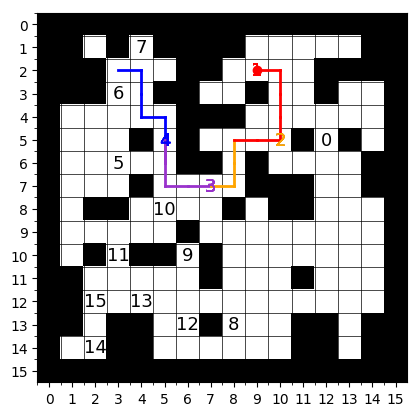}}
&\subfloat[]{\includegraphics[width=0.47\textwidth]{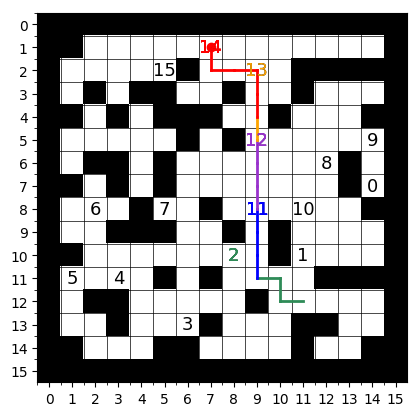}}\\
\subfloat[]{\includegraphics[width=0.47\textwidth]{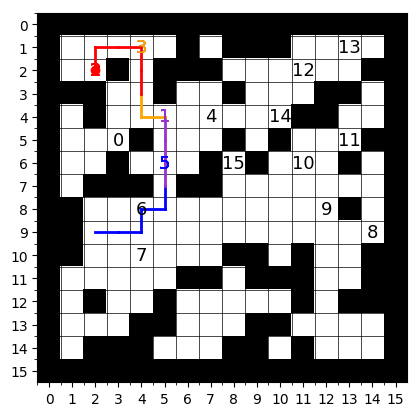}}
&\subfloat[]{\includegraphics[width=0.47\textwidth]{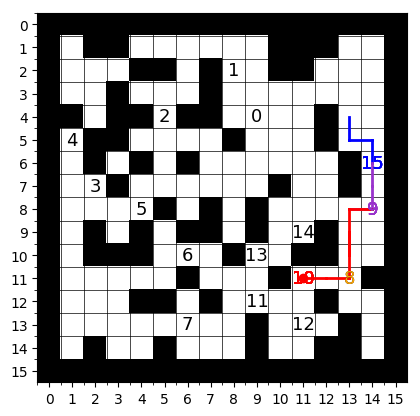}}
\end{tabular}
\caption{Trajectories generated by our method on the unseen grid-world maps for both the seen goals (a) (b) and the unseen goals (c) (d). The different colors represent different sub-goals and the corresponding sub-goal-oriented trajectories where the red one denotes the designated final goal. }
\label{fig:gw_results}
\end{figure*}

\section{Experiments of the Robotic Object Search}
\label{appendix:ros}

\subsection{Experiments on AI2-THOR \cite{kolve2017ai2}}

\begin{figure*}[ht]
\centering
\begin{tabular}{cc}
\includegraphics[width=0.45\textwidth]{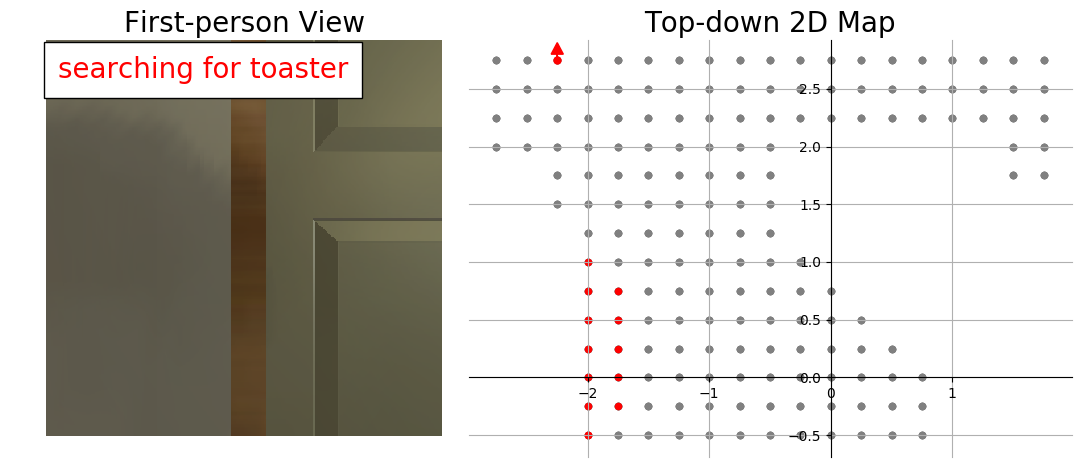} &
\includegraphics[width=0.45\textwidth]{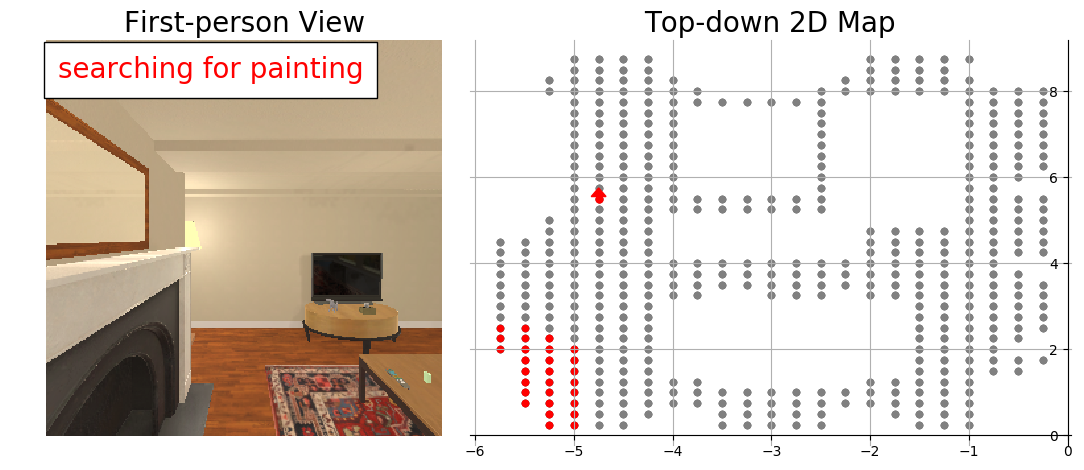} \\
\contour{black}{$\Downarrow$} & \contour{black}{$\Downarrow$} \\
\includegraphics[width=0.45\textwidth]{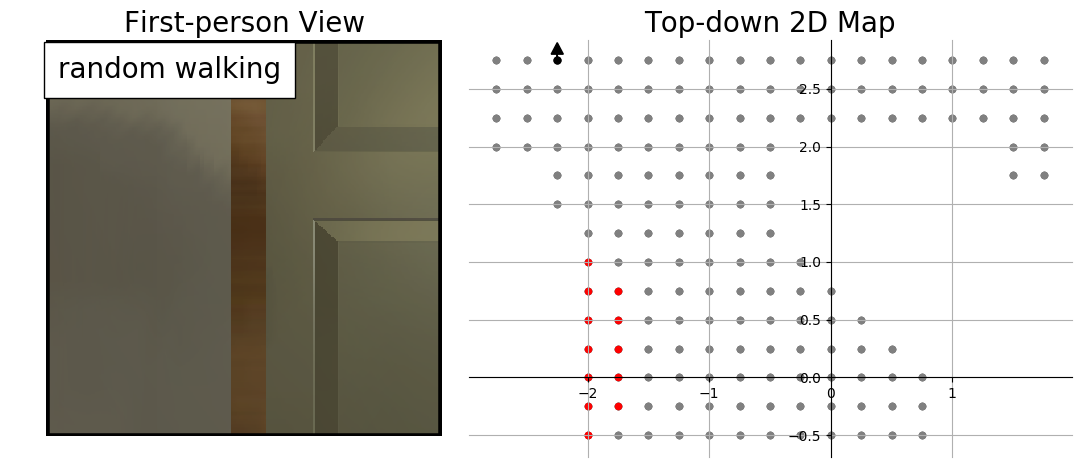} &
\includegraphics[width=0.45\textwidth]{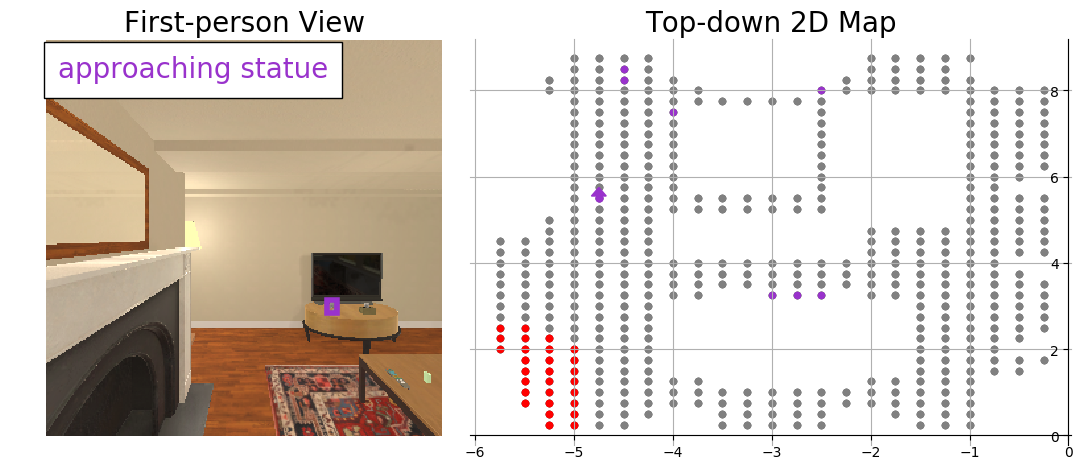} \\
\contour{black}{$\Downarrow$} & \contour{black}{$\Downarrow$} \\
\includegraphics[width=0.45\textwidth]{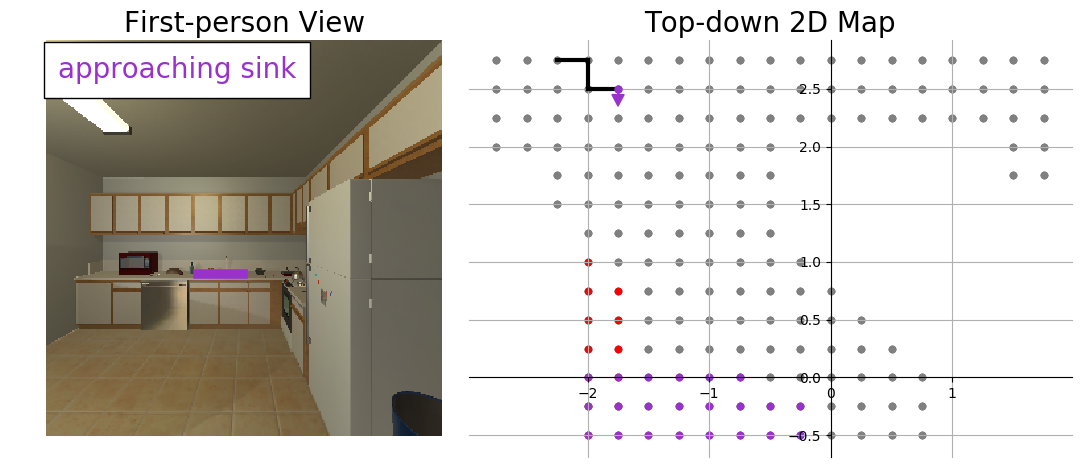} &
\includegraphics[width=0.45\textwidth]{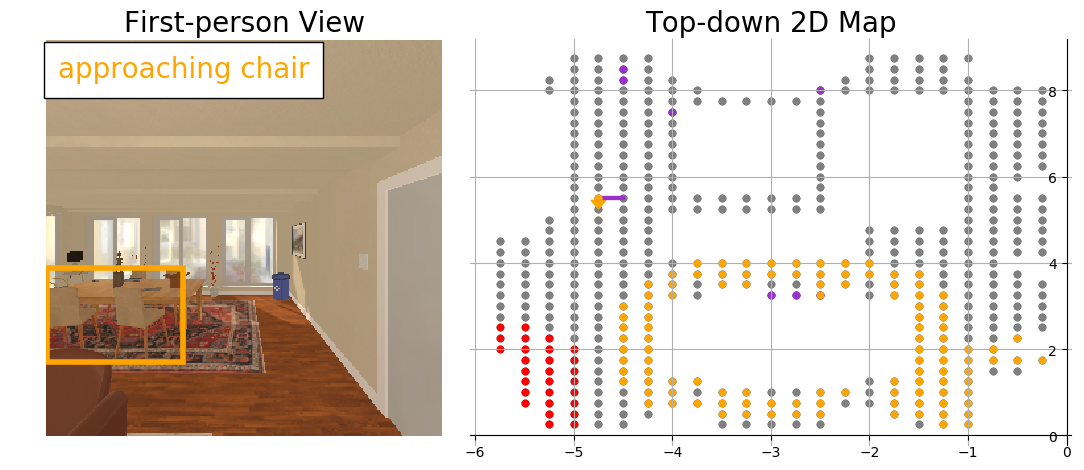} \\
\contour{black}{$\Downarrow$} & \contour{black}{$\Downarrow$} \\
\includegraphics[width=0.45\textwidth]{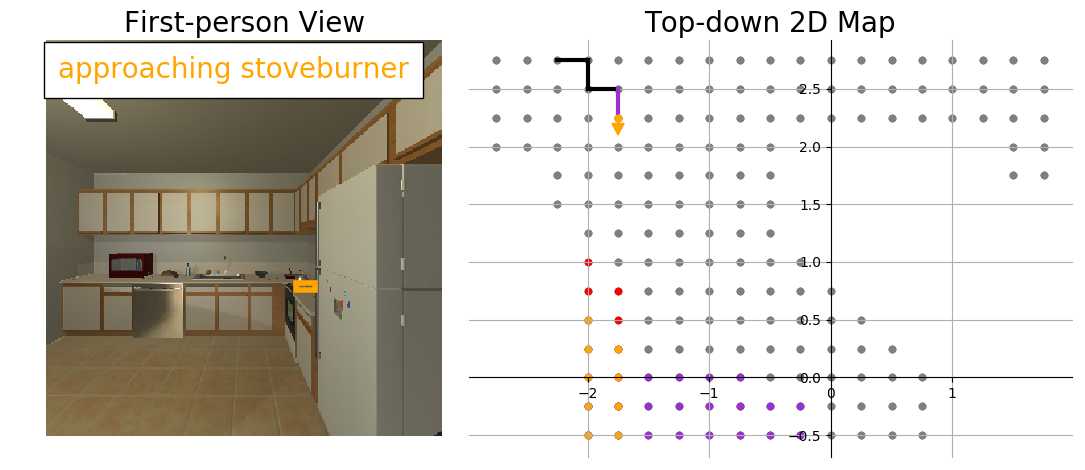} &
\includegraphics[width=0.45\textwidth]{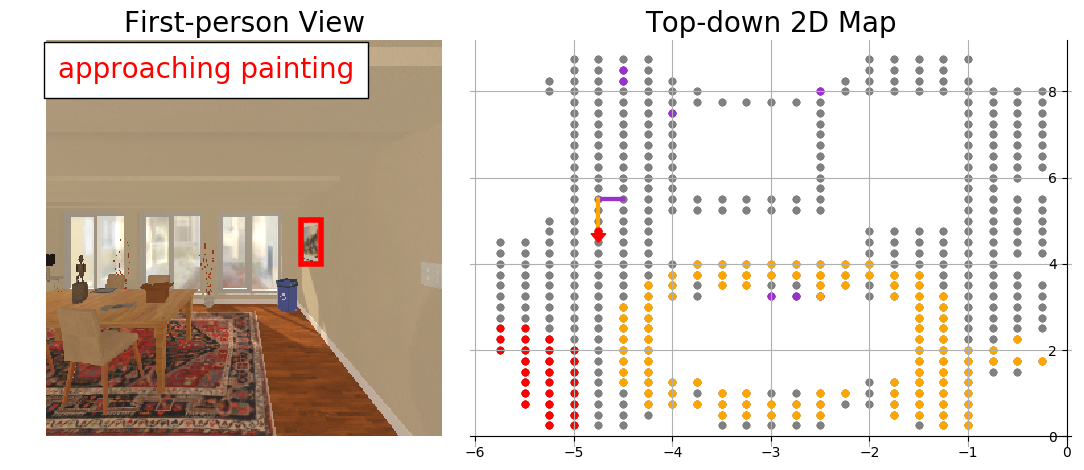} \\
\contour{black}{$\Downarrow$} & \contour{black}{$\Downarrow$} \\
\includegraphics[width=0.45\textwidth]{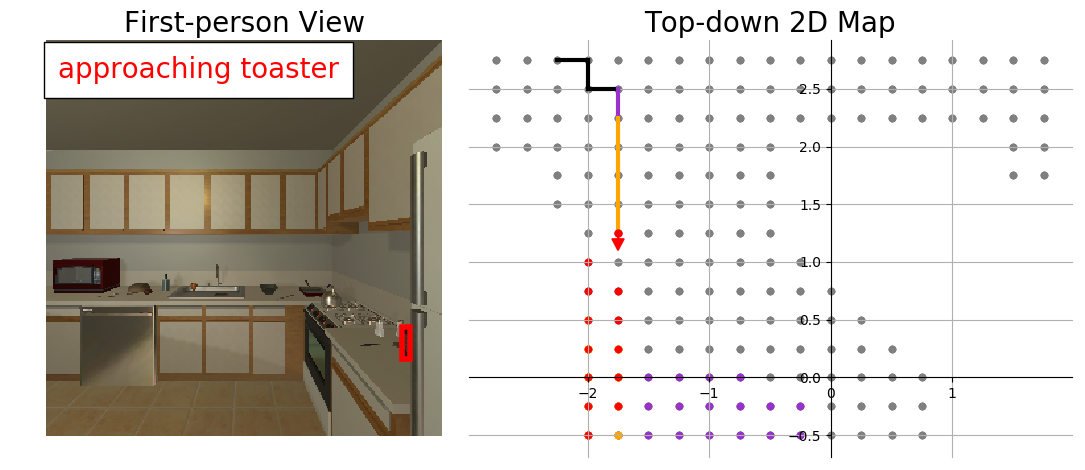} &
\includegraphics[width=0.45\textwidth]{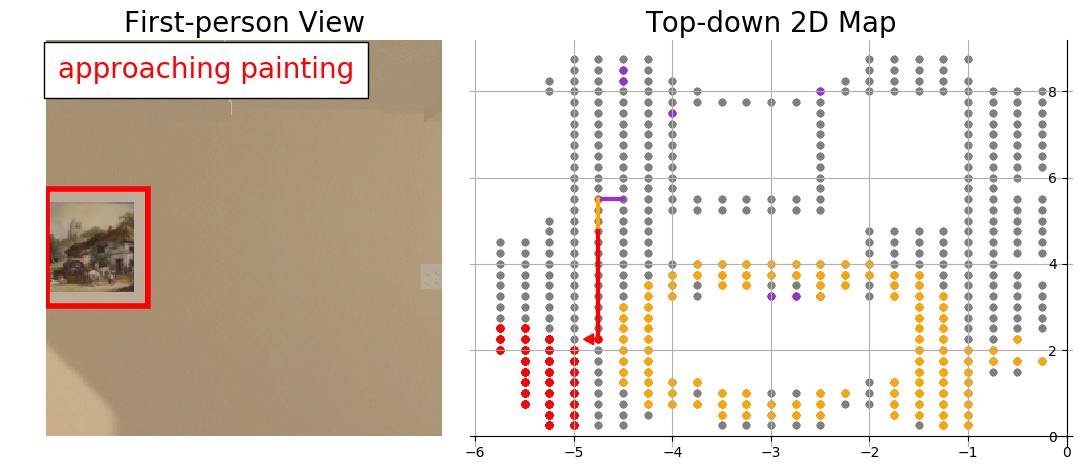} \\
\contour{black}{$\Downarrow$} & \contour{black}{$\Downarrow$} \\
\includegraphics[width=0.45\textwidth]{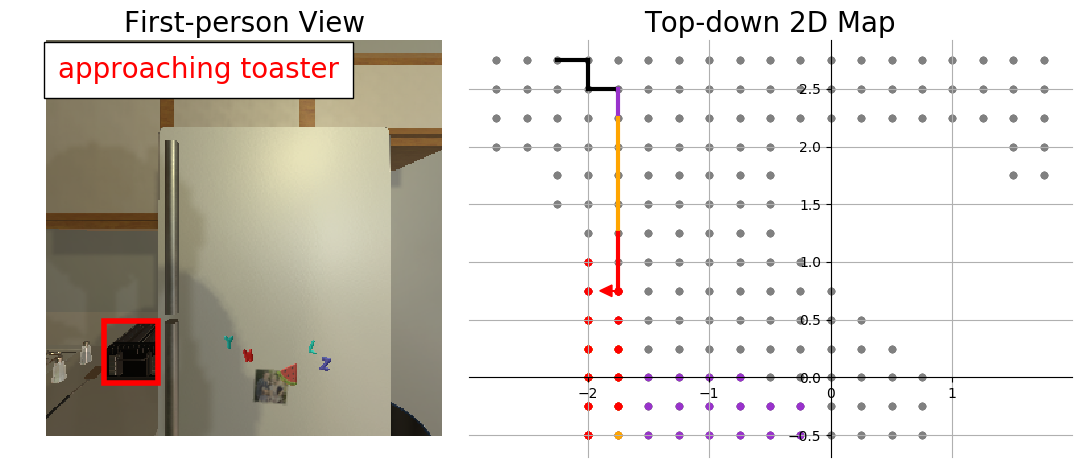} &
\includegraphics[width=0.45\textwidth]{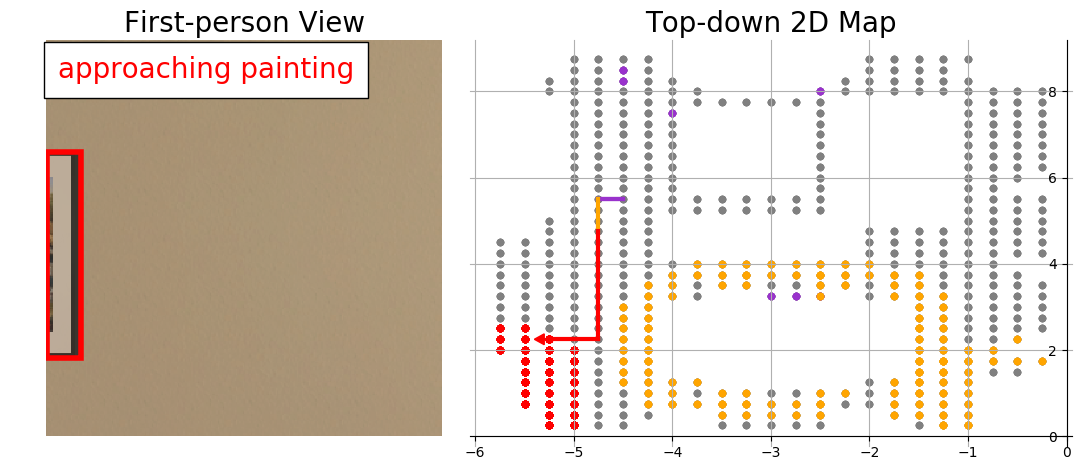} \\
(a) kitchen (toaster) & (b) living room (painting) \\
\end{tabular}
% \caption{}
% \label{fig:qua_results}
\end{figure*}

\begin{figure*}[ht]
\centering
\begin{tabular}{cc}
\includegraphics[width=0.45\textwidth]{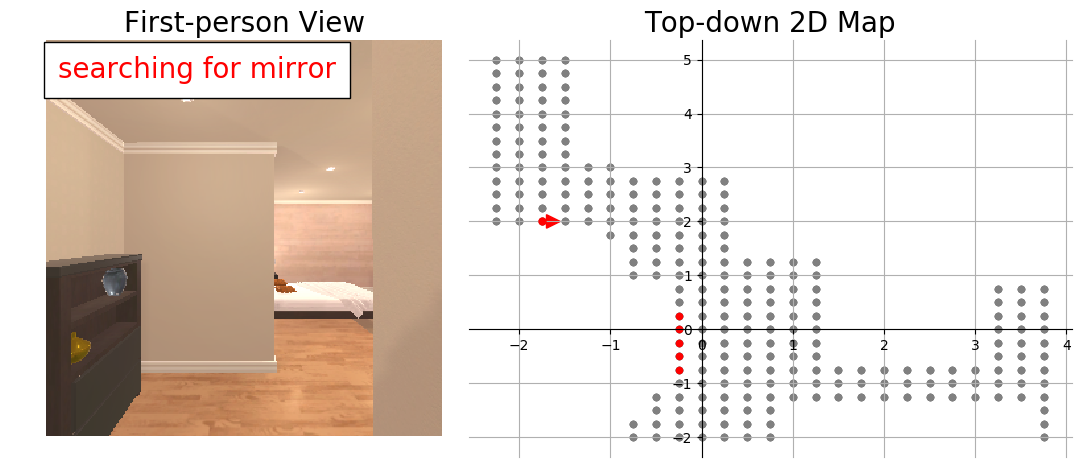} &
\includegraphics[width=0.45\textwidth]{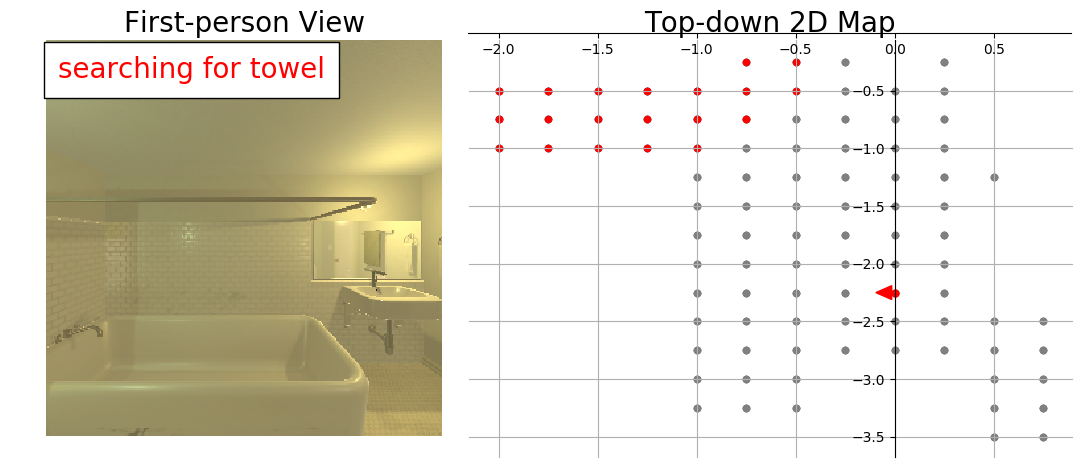} \\
\contour{black}{$\Downarrow$} & \contour{black}{$\Downarrow$} \\
\includegraphics[width=0.45\textwidth]{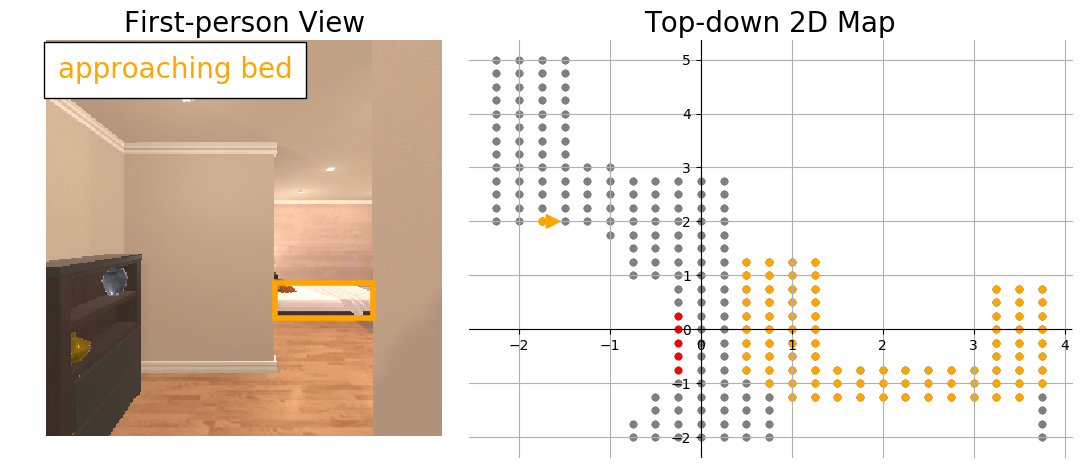} &
\includegraphics[width=0.45\textwidth]{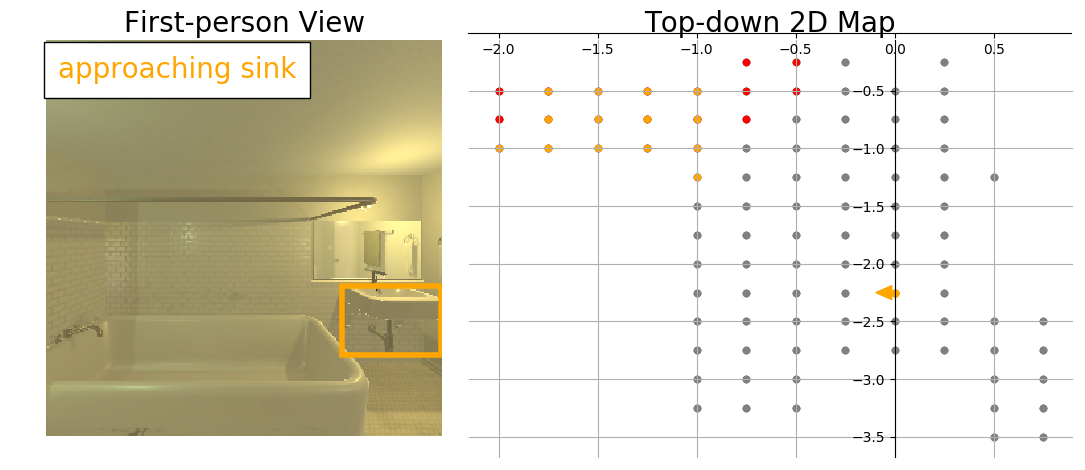} \\
\contour{black}{$\Downarrow$} & \contour{black}{$\Downarrow$} \\
\includegraphics[width=0.45\textwidth]{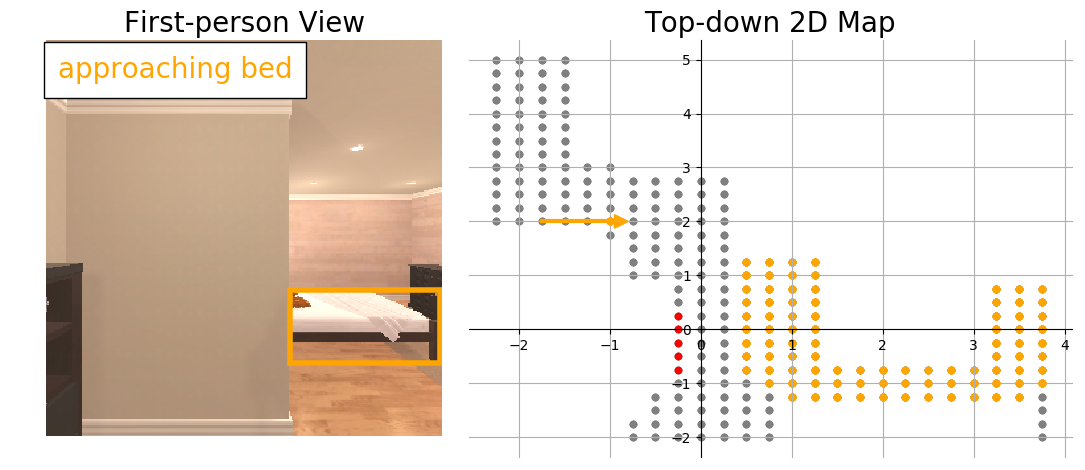} &
\includegraphics[width=0.45\textwidth]{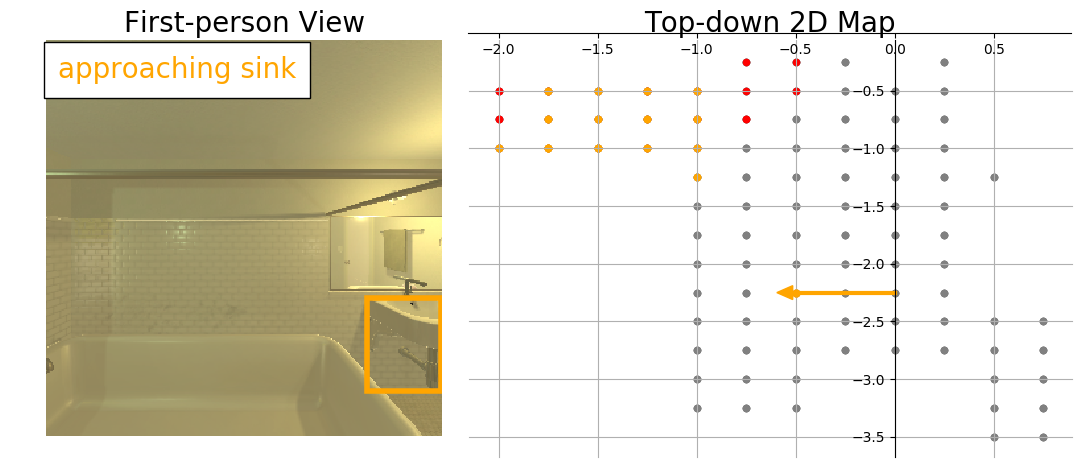} \\
\contour{black}{$\Downarrow$} & \contour{black}{$\Downarrow$} \\
\includegraphics[width=0.45\textwidth]{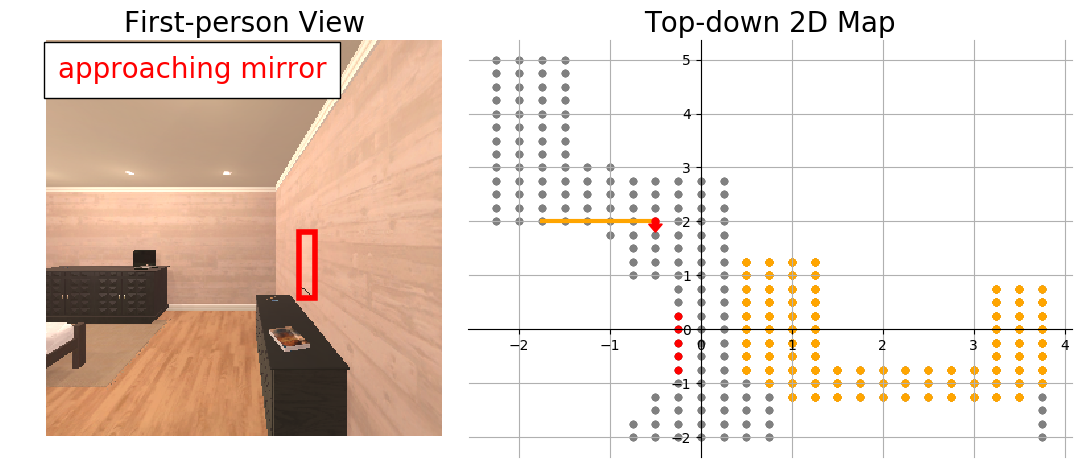} &
\includegraphics[width=0.45\textwidth]{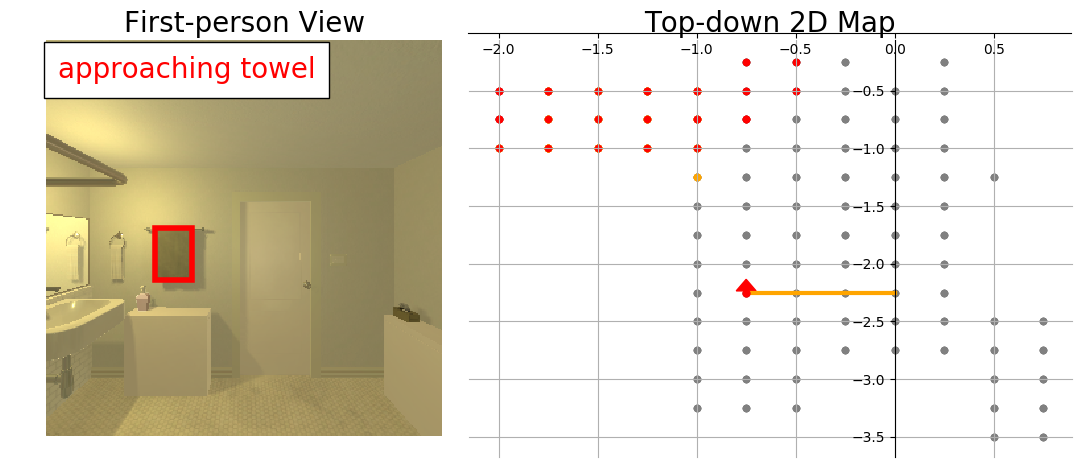} \\
\contour{black}{$\Downarrow$} & \contour{black}{$\Downarrow$} \\
\includegraphics[width=0.45\textwidth]{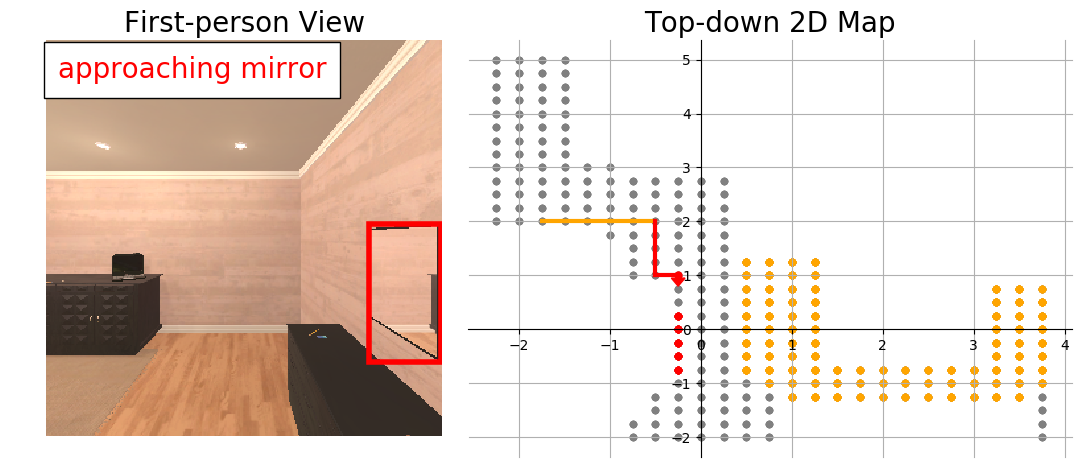} &
\includegraphics[width=0.45\textwidth]{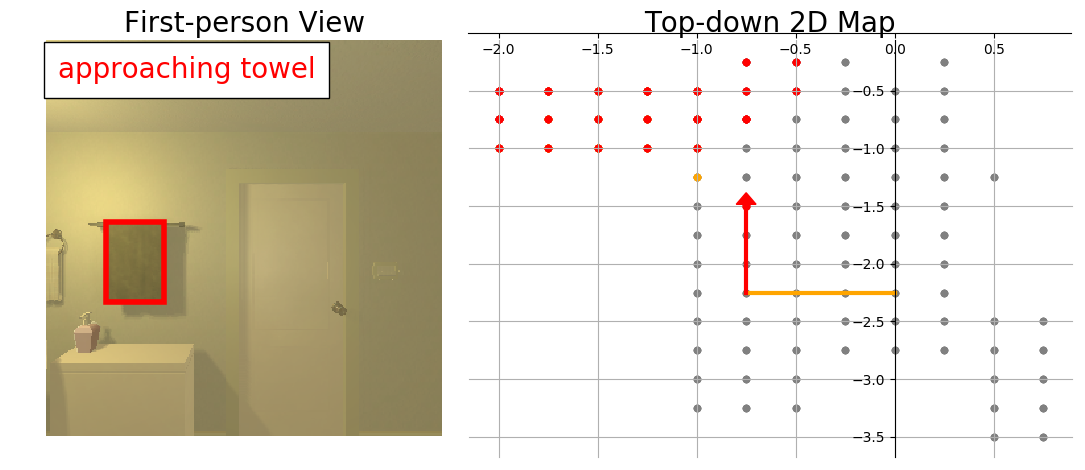} \\
\contour{black}{$\Downarrow$} & \contour{black}{$\Downarrow$} \\
\includegraphics[width=0.45\textwidth]{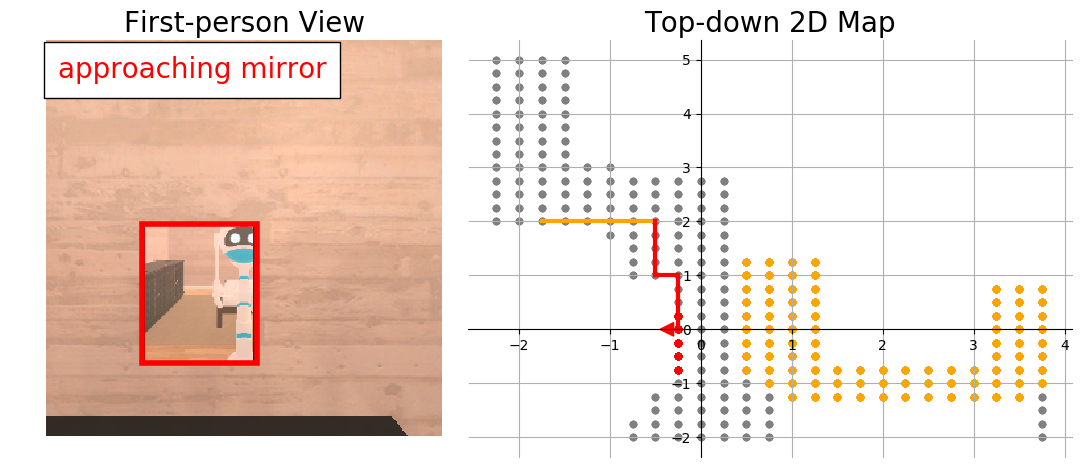} &
\includegraphics[width=0.45\textwidth]{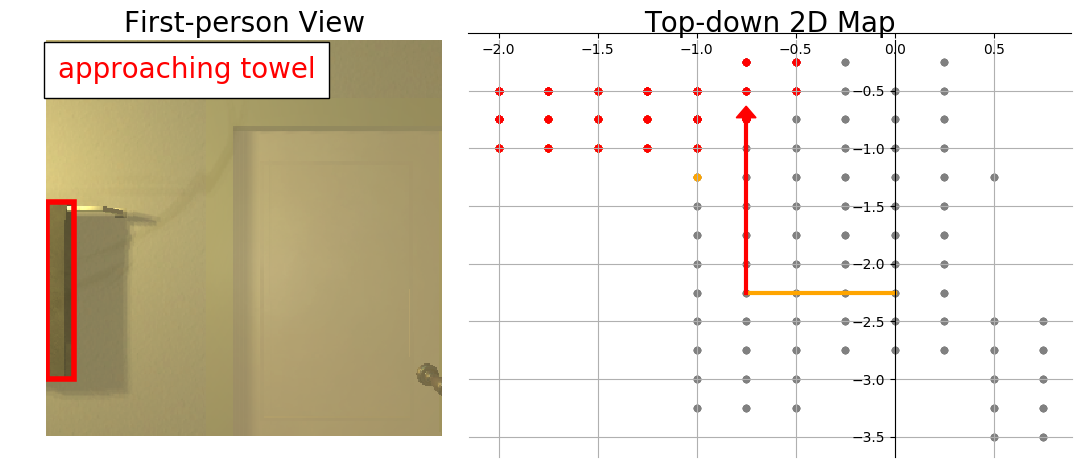} \\
(c) bedroom (mirror) & (d) bathroom (towel) \\
\end{tabular}
\caption{Trajectories generated by our method for the robotic object search task on AI2-THOR \cite{kolve2017ai2}.}
\label{fig:ros_ai2thor_results}
\end{figure*}

\subsubsection{Network Architecture}
We take the first-person view semantic segmentation and the depth map of window size $30 \times 30$ as the agent's pre-processed observation. We concatenate $4$ history observations to input to our model. Specifically, the high-level network of our method \textsc{HRL-GRG} takes a sub-goal specified semantic segmentation of size $4 \times 30 \times 30 \times 1$ and the depth map of the same size as the inputs. The semantic segmentation and the depth map are first concatenated to size $4 \times 30 \times 30 \times 2$ and then fed into two convolutional layers with $50$ and $100$ filters respectively of kernel size $3 \times 3$ where a $2 \times 2$ max-pooling is attached after each layer. The results further pass through two fully-connected layers with $256$ hidden units and $1$ output respectively. The output is the predicted Q-value of the input sub-goal. 

The low-level network of our method \textsc{HRL-GRG} takes the same inputs as the high-level network. For each input stream, our low-level network first flattens it and project it to a $256$ dimensional vector. The two $256$ dimensional vectors are then concatenated and projected to a joint $256$ dimensional vector before passing through two branches. Each branch has two fully-connected layers with $20$ hidden units in the first layer. The first branch outputs a $4$ dimensional vector which is further converted to a probability distribution over the $4$ actions using the softmax function, and the second branch outputs a single state value. All the hidden fully-connected layers are activated by the ReLU function.

\subsubsection{Training Protocols and Hyperparameters}
We follow the same experimental setting in \cite{yang2018visual} (without ``stop'' action). In addition, we adopt the Double DQN \cite{van2016deep} technique for our high-level network and we pre-train our low-level network to approach a visible object. We train our method with the curriculum training paradigm we described in Section~\ref{appendix:gw_train}. All hyperparameters are summarized in Table~\ref{tbl:ros_hp} and the detailed results are reported in Table~\ref{tbl:ros_ai2thor_full}.

\begin{table*}
\caption{The performance of \textsc{Scene Priors} \cite{yang2018visual} (without stop action) and our \textsc{HRL-GRG} in performing robotic object search task on AI2-THOR \cite{kolve2017ai2}. (A: performance improvement; B and C: performance of the method and the \textsc{Random} method.)}
    \centering
    \begin{tabular}{ccccccc}
    \multicolumn{2}{c}{\multirow{2}{*}{+A (B - C)}} & \multicolumn{2}{c}{Seen Goals}&& \multicolumn{2}{c}{Unseen Goals}\\
    \cline{3-4}\cline{6-7}
    & & SR$\uparrow$ &SPL$\uparrow$ & &SR$\uparrow$ & SPL$\uparrow$\\
    \specialrule{0.12em}{1pt}{1.5pt}
    \multirow{2}{*}{Seen Env.} &  \textsc{Scene Priors}   &+0.25 (0.62 - 0.37) &+0.16 (0.26 - 0.10) && +0.08 (0.48 - 0.40) &+0.07 (0.18 - 0.11)\\
    & \textbf{HRL-GRG} &\textbf{+0.37} (0.74 - 0.37) &\textbf{+0.24} (0.34 - 0.10) &&\textbf{+0.33} (0.73 - 0.40) &\textbf{+0.23} (0.34 - 0.11)\\
    \specialrule{0.01em}{1pt}{1pt}
    \multirow{2}{*}{Unseen Env.} &   \textsc{Scene Priors} &+0.18 (0.56 - 0.38) &+0.11 (0.21 - 0.10) && +0.12 (0.49 - 0.37) &+0.06 (0.16 - 0.10) \\
   & \textbf{HRL-GRG} &\textbf{+0.33} (0.71 - 0.38) &\textbf{+0.21} (0.31 - 0.10) &&\textbf{+0.38} (0.75 - 0.37) &\textbf{+0.23} (0.33 - 0.10)\\
   \specialrule{0.12em}{1pt}{1.5pt}
    \end{tabular}
    \label{tbl:ros_ai2thor_full}
\end{table*}

\subsubsection{Qualitative Results}
Figure~\ref{fig:ros_ai2thor_results} shows some examples of how our method searches for unseen objects in unseen AI2-THOR \cite{kolve2017ai2} scenes. The results shall be better viewed in the supplementary videos.

\subsection{Experiments on House3D \cite{wu2018building}}
\subsubsection{Data Pre-processing}
% We conduct the robotic object search task on the House3D \cite{wu2018building} environment. 
In House3D simulation, the environments and goals we adopt for both the single environment setting and the multiple environments setting are shown in Table~\ref{tbl:ros_env}. For each environment, we consider discrete actions for the agent to navigate. Specifically, the agent moves forward / backward / left / right $0.2$ meters, or rotates $90$ degrees at each time step, which discretizes the environment into a certain number of reachable locations. We collect the first-person view RGB images as well as the corresponding the ground-truth semantic segmentations and depth maps at every locations of $100$ preserved environments (other than those shown in Table~\ref{tbl:ros_env}) as the training data to train the encoder-decoder model \cite{chen2018encoder} to predict both the semantic segmentations and the depth maps from the first-person RGB images. For our robotic object search task, we resize the both predictions to the size $10 \times 10$ and take them as the agent's observation. We consider an object that is unique in an environment as a valid target object for the agent to search, and we define the goal locations as the $5$ locations that yields the observations with the largest target object area. 

\newcommand{\minitab}[2][l]{\begin{tabular}{#1}#2\end{tabular}}
\begin{table*}
    \small
    \caption{The environments and goals we adopt for the robotic object search task on House3D \cite{wu2018building} .}
    \label{tbl:ros_env}
    \centering
    \begin{tabular}{c|c|l|c|l}
        \hline
         \multirow{4}{*}{\minitab[c]{Single \\ Environment}} & \multirow{4}{*}{\minitab[c]{Seen \\ Env}} &\multirow{4}{*}{5cf0e1e9493994e483e985c436b9d3bc} & \multirow{2}{*}{\minitab[c]{Seen \\ Goals}} & music, television, heater,\\
         & & & & stand, dressing table, table \\
         \cline{4-5}
         & & & \multirow{2}{*}{\minitab[c]{Unseen \\ Goals}} & bed, mirror, ottoman,\\
         & & &  & sofa, desk, picture frame \\
         \hline
         \hline
         \multirow{8}{*}{\minitab[c]{Multiple \\ Environments}} & \multirow{4}{*}{\minitab[c]{Seen \\ Envs}} & 5cf0e1e9493994e483e985c436b9d3bc & \multirow{4}{*}{Goals} & \multirow{4}{*}{\minitab[l]{sofa, bed, television, \\ tv stand, toilet, bathtub}}\\
         & &0c9a666391cc08db7d6ca1a926183a76&& \\
         & & 0c90efff2ab302c6f31add26cd698bea& & \\
         & &00d9be7210856e638fa3b1addf2237d6 && \\
         \cline{2-5}
          & \multirow{4}{*}{\minitab[c]{Unseen \\ Envs}} & 07d1d46444ca33d50fbcb5dc12d7c103&\multirow{4}{*}{Goals} & \multirow{4}{*}{\minitab[l]{ sofa, bed, dressing table, \\ mirror, ottoman, music}}\\
         & &026c1bca121239a15581f32eb27f2078 && \\
         & & 0147a1cce83b6089e395038bb57673e3& & \\
         & & 0880799c157b4dff08f90db221d7f884 && \\
         \hline
    \end{tabular}
\end{table*}

\subsubsection{Network Architectures}
For all the methods, we concatenate $4$ history observations for the agent to make a decision. We describe the architectures of all methods in details as follows.
\begin{itemize}
    \item \textsc{DQN}. The vanilla DQN implementation that takes the semantic segmentation of size $4 \times 10 \times10 \times 78$ and the depth map of size $4 \times 10 \times 10 \times 1$ as the inputs. The segmentation input is first passed through a convolutional layer with $1$ filter of kernel size $1\times 1$ to reduce its channel size to $1$. Then for both the segmentation input and the depth input, we flatten each of which to a $400$ dimensional vector and project it down to a $256$ dimensional vector through a fully-connected layer respectively. The two $256$ dimensional vectors, as well as the $78$ dimensional one-hot vector representing the target object are further concatenated and fed into two fully-connected layers with $256$ hidden units and $6$ outputs as the Q-values of $6$ actions. Each hidden fully-connected layer is activated by the ReLU function.
    \item \textsc{A3C}. The vanilla A3C implementation takes the target object specified channel of the semantic segmentation that has size $4 \times 10 \times 10 \times 1$ and the depth map of size $4 \times 10 \times 10 \times 1$ as the inputs. For each input stream, we flatten it and project it to a $256$ dimensional vector. The two $256$ dimensional vectors are further concatenated and passed through two branches. Each branch has two fully-connected layers with $20$ hidden units in the first layer. The first branch further projects the $20$ dimensional vector to $6$ outputs which are converted to probabilities of $6$ actions using the softmax function, and the second branch outputs a single state value. All the hidden fully-connected layers are activated by the ReLU function.
    \item \textsc{Hrl}. The high-level network of \textsc{Hrl} is similar to the method \textsc{DQN} while it outputs $78$ values representing the Q-values of the $78$ sub-goals. The low-level network is exactly same as the method \textsc{A3C}.
    \item \textsc{Ours}. The high-level network of our method \textsc{HRL-GRG} is similar to the method \textsc{A3C} while the first branch is removed and only the second branch is in place to output a Q-value of the input sub-goal. The low-level network is exactly same as the method \textsc{A3C}.
    
\end{itemize}

\subsubsection{Training Protocols and Hyperparameters}
Similar to that in the grid-world domain, we also adopt the Double DQN \cite{van2016deep} technique for all the DQN networks. We pre-train the method \textsc{A3C} to approach an object when the object is observable and we take it as the pre-trained model for the low-level networks of both \textsc{Hrl} and our method as well. For all the methods, we adopt the same curriculum training paradigm as we described in Section~\ref{appendix:gw_train} and we summarize the hyperparameters in Table~\ref{tbl:ros_hp} .

\begin{table*}
    % \small
    \caption{Hyperparameters of all the methods for the robotic object search task.}
    \label{tbl:ros_hp}
    \centering
    \begin{tabular}{lll}
        \specialrule{0.12em}{0pt}{2pt}
        Hyperparameter & Description & Value\\
         \specialrule{0.05em}{0pt}{2pt}
         $\gamma$ & Discount factor & 0.99\\
         lr & Learning rate for all networks (high-level/low-level) & 0.0001\\
         main\_update & Interval of updating all the main networks of Double DQN & 100 \\
         target\_update & Interval of updating all the target networks of Double DQN & 100000\\
         A3C\_update & Interval of updating all the A3C networks & 10\\
         $\beta$ & The weight of the entropy regularization term in the A3C networks & 0.01 \\
         batch\_size & Batch size for training DQN networks & 64 \\
         epsilon & Initial exploration rate, anneal episodes, final exploration rate & 1, 10000, 0.1 \\ 
         max\_episodes & Maximum episodes to train each method & 100000\\
         $N_{max}^l$ & The maximum steps that low-level network can take in AI2-THOR / House3D & 10 / 50 \\
         $N_{max}$ & The maximum steps that each method can take in AI2-THOR / House3D & \cite{yang2018visual} / 1000 \\
         optimizer & Optimizer for all the networks & RMSProp \\
         $\bm{\alpha_{ij}}$ & The hyperparameter of our GRG $(\alpha_{ij,1},...,\alpha_{ij,10},\alpha_{i,j,11})$ & $(0,...,0,1)$\\
         \specialrule{0.12em}{0pt}{2pt}
    \end{tabular}
\end{table*}

\subsubsection{Qualitative Results}
\begin{figure*}[ht]
\centering
\begin{tabular}{cc}
\includegraphics[width=0.48\textwidth]{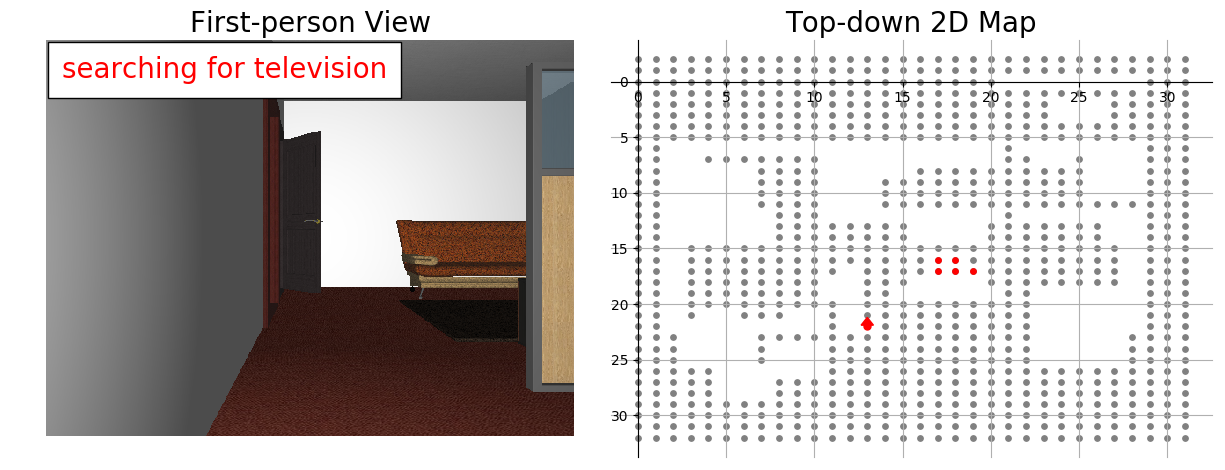} &
\includegraphics[width=0.48\textwidth]{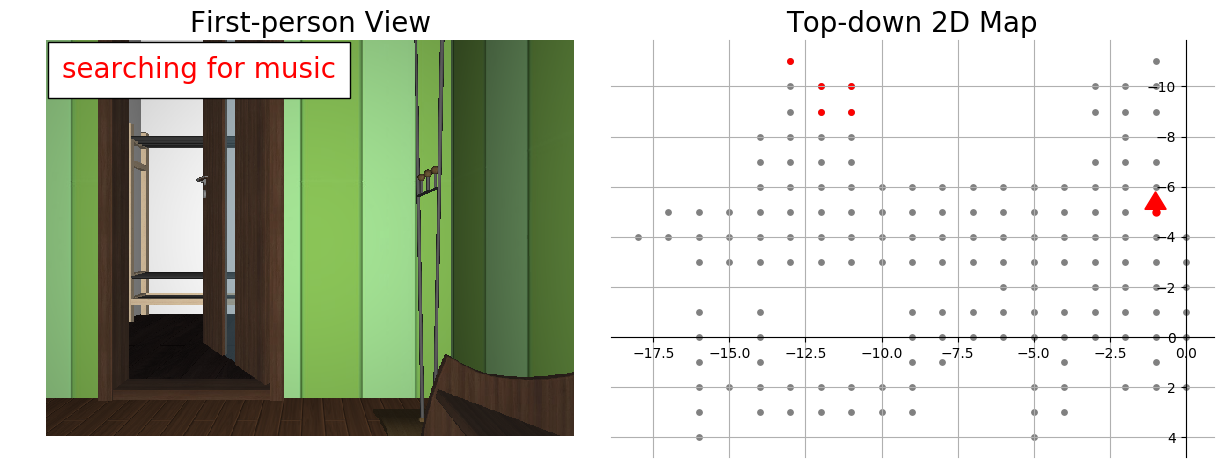} \\
\contour{black}{$\Downarrow$} & \contour{black}{$\Downarrow$} \\
\includegraphics[width=0.48\textwidth]{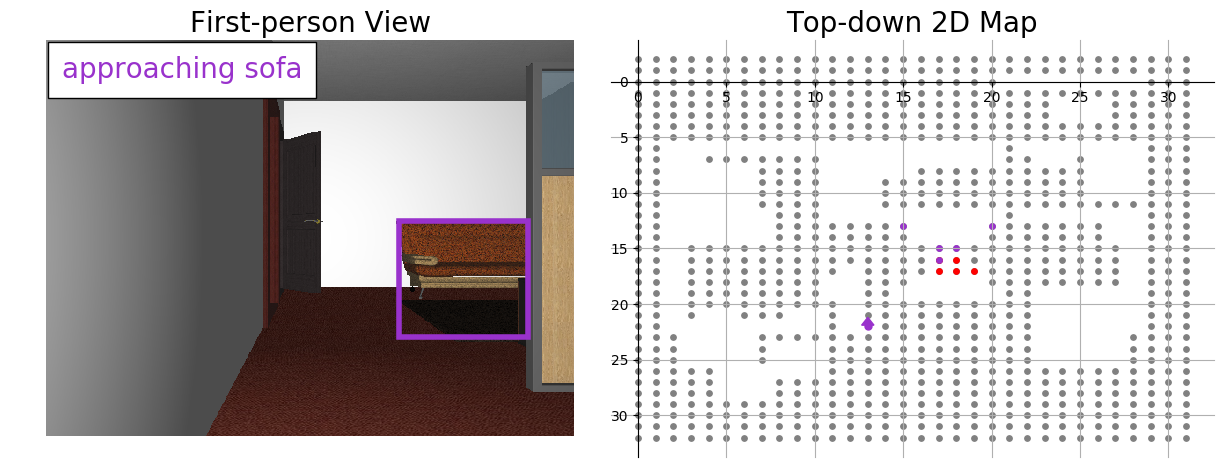} &
\includegraphics[width=0.48\textwidth]{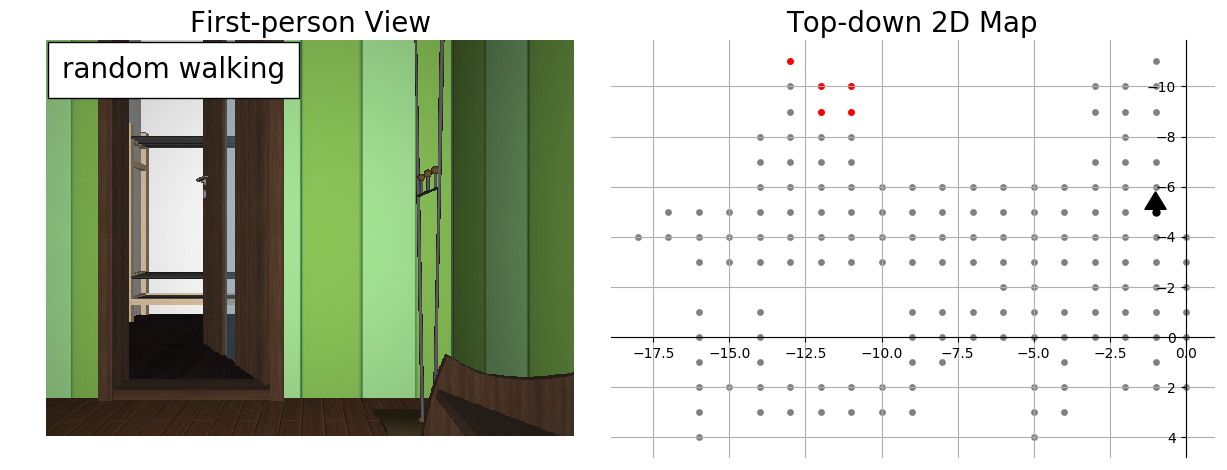} \\
\contour{black}{$\Downarrow$} & \contour{black}{$\Downarrow$} \\
\includegraphics[width=0.48\textwidth]{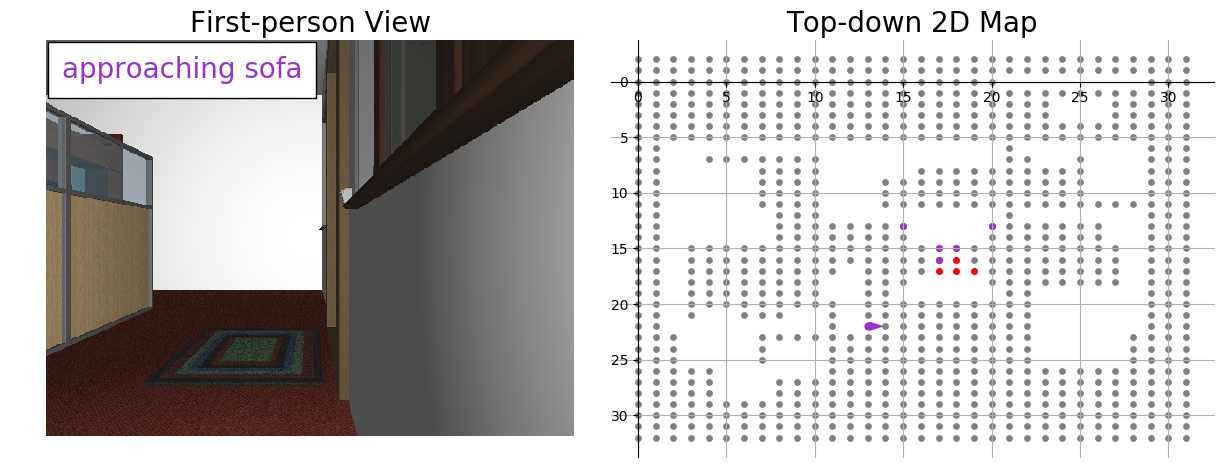} &
\includegraphics[width=0.48\textwidth]{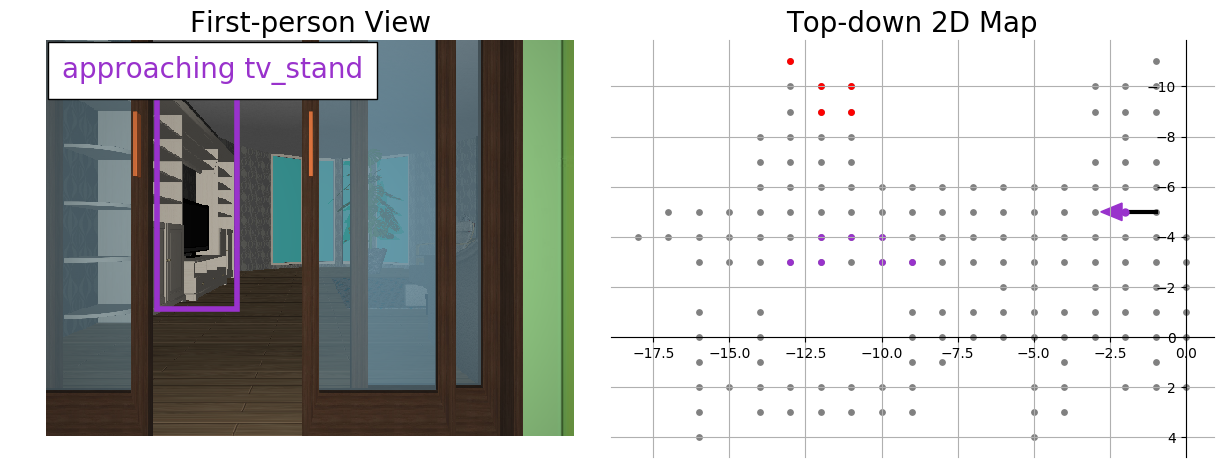} \\
\contour{black}{$\Downarrow$} & \contour{black}{$\Downarrow$} \\
\includegraphics[width=0.48\textwidth]{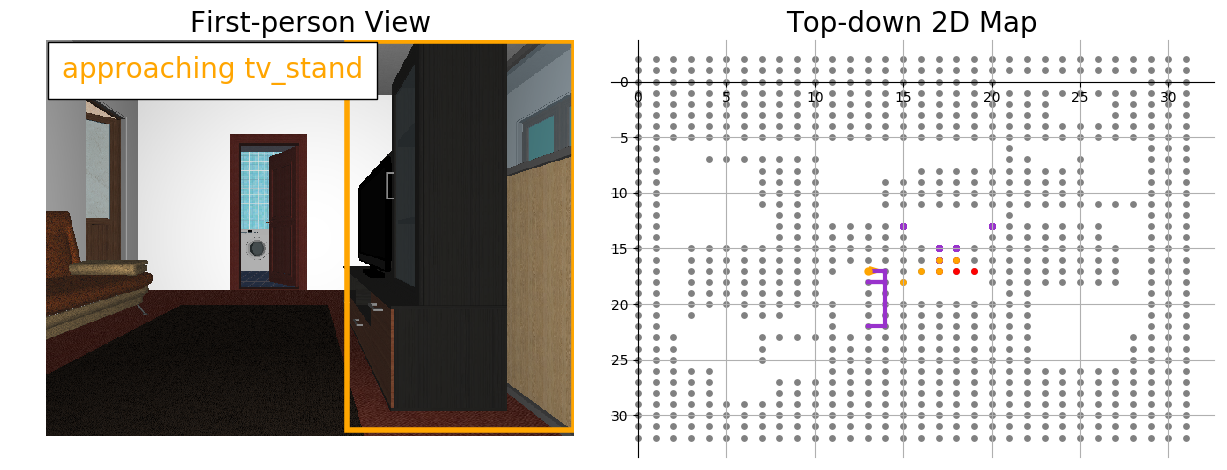} &
\includegraphics[width=0.48\textwidth]{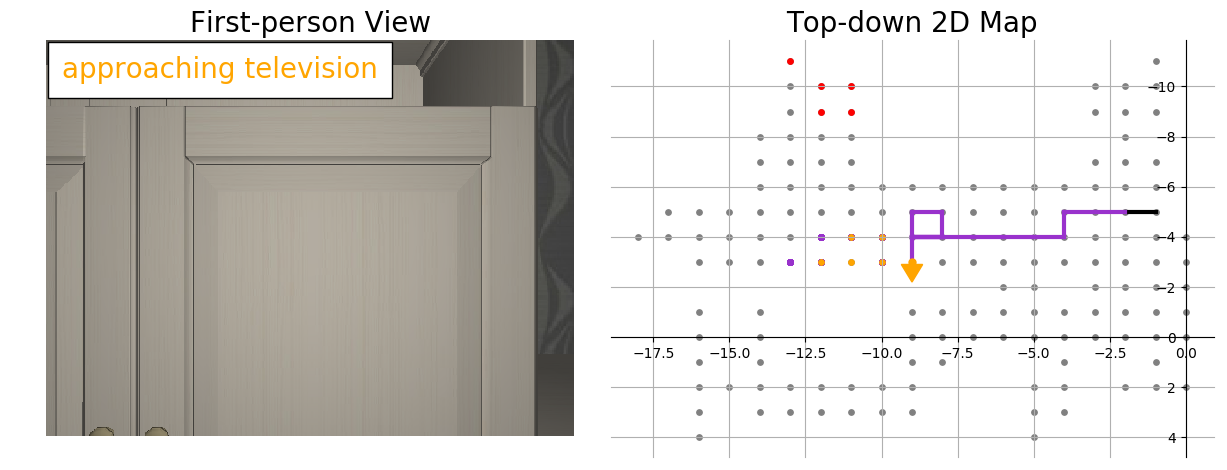} \\
\contour{black}{$\Downarrow$} & \contour{black}{$\Downarrow$} \\
\includegraphics[width=0.48\textwidth]{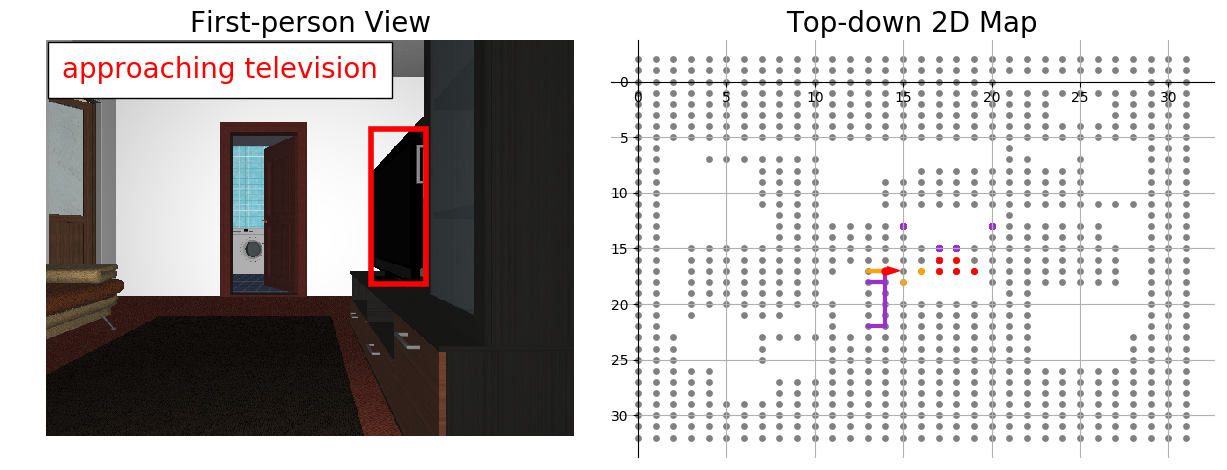} &
\includegraphics[width=0.48\textwidth]{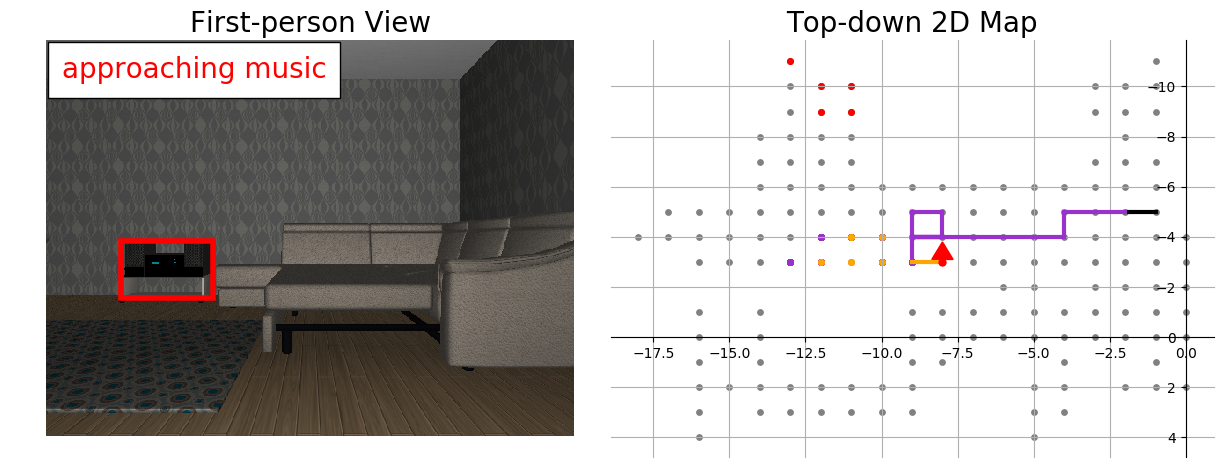} \\
\contour{black}{$\Downarrow$} & \contour{black}{$\Downarrow$} \\
\includegraphics[width=0.48\textwidth]{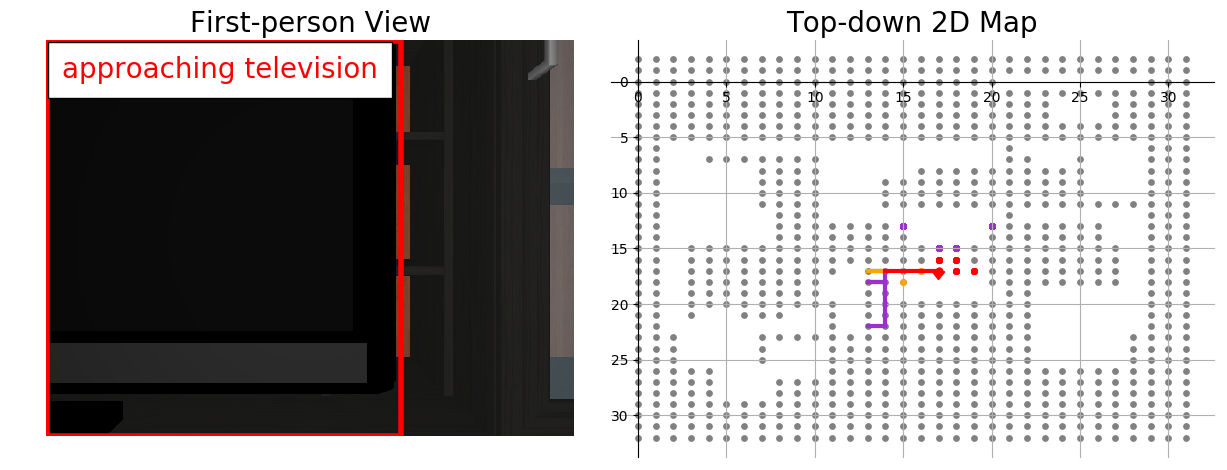} &
\includegraphics[width=0.48\textwidth]{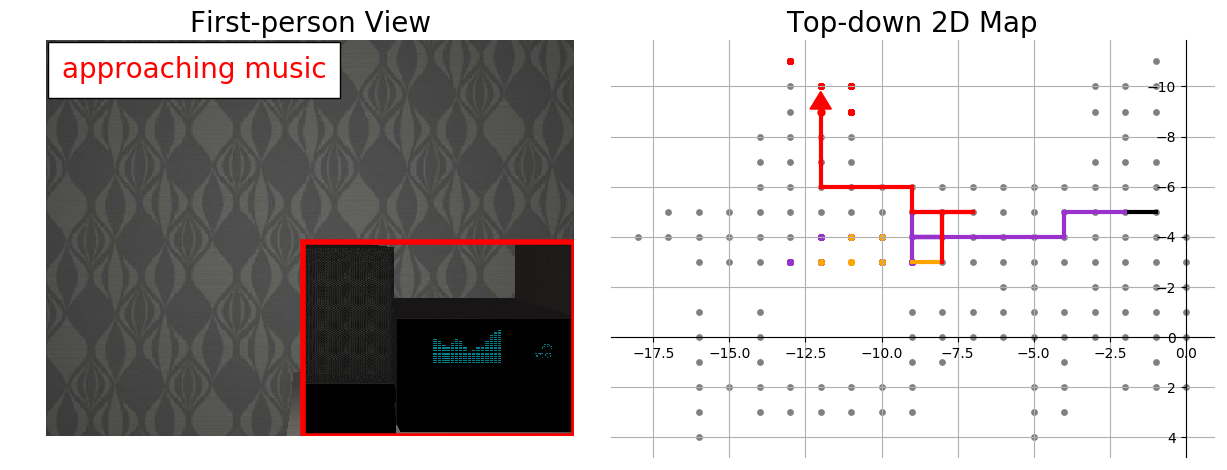} \\
(c) unseen environment seen goal & (d) unseen environment unseen goal \\
\end{tabular}
\caption{Trajectories generated by our method for the robotic object search task on House3D \cite{wu2018building}.}
\label{fig:ros_house3d_results}
\end{figure*}

Figure~\ref{fig:ros_house3d_results} shows some qualitative results of our method for the robotic object search task on House3D \cite{wu2018building}. The agent can only access the first-person view RGB images while the top-down 2D maps are placed for better visualization. The results shall be better viewed in the supplementary videos.
\begin{figure*}[ht]
\centering
\begin{tabular}{cc}
\includegraphics[width=0.48\textwidth]{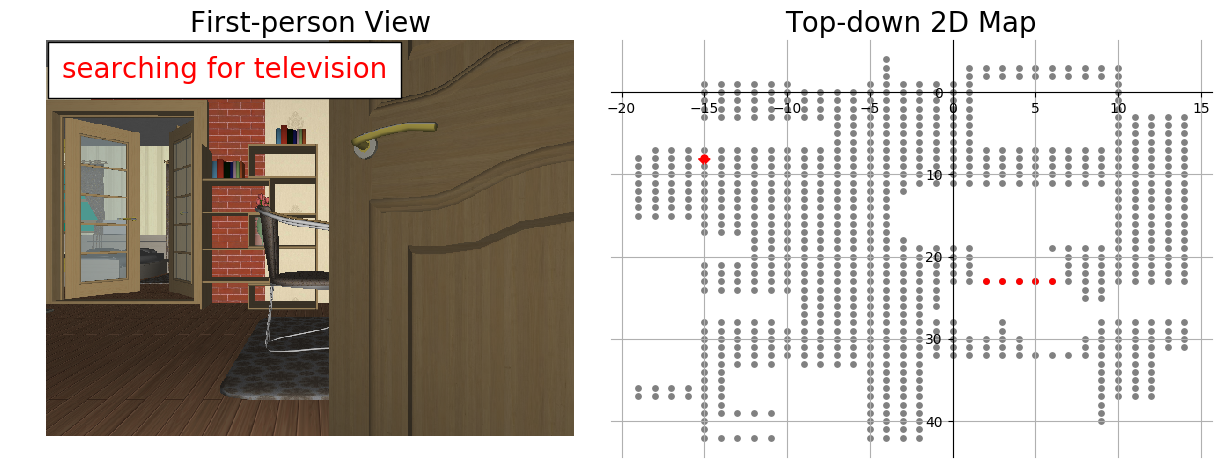} &
\includegraphics[width=0.48\textwidth]{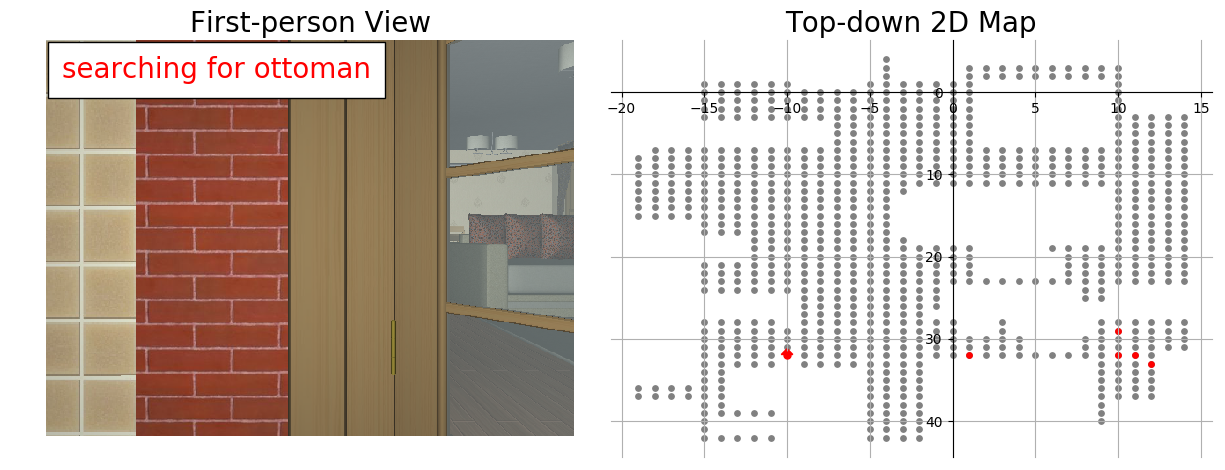} \\
\contour{black}{$\Downarrow$} & \contour{black}{$\Downarrow$} \\
\includegraphics[width=0.48\textwidth]{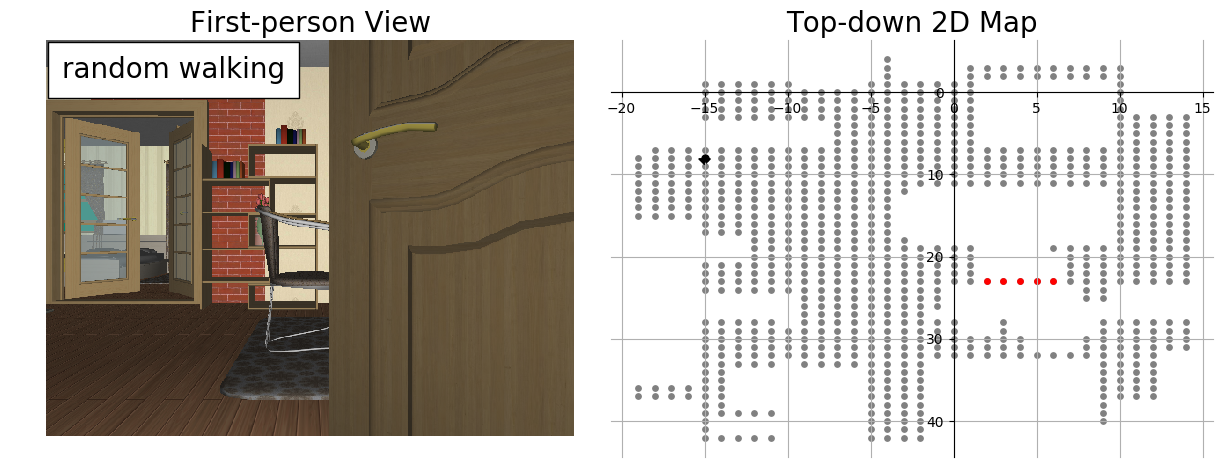} &
\includegraphics[width=0.48\textwidth]{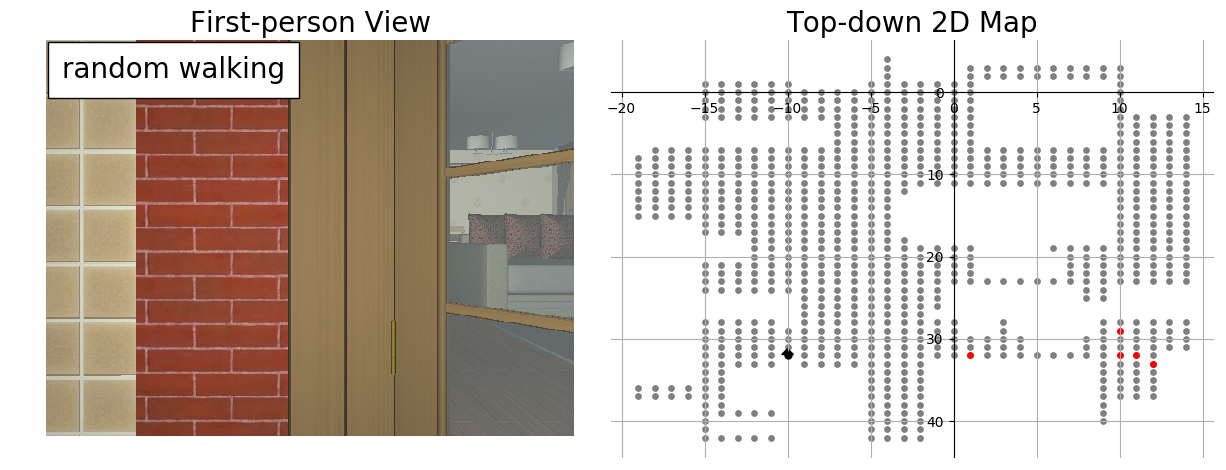} \\
\contour{black}{$\Downarrow$} & \contour{black}{$\Downarrow$} \\
\includegraphics[width=0.48\textwidth]{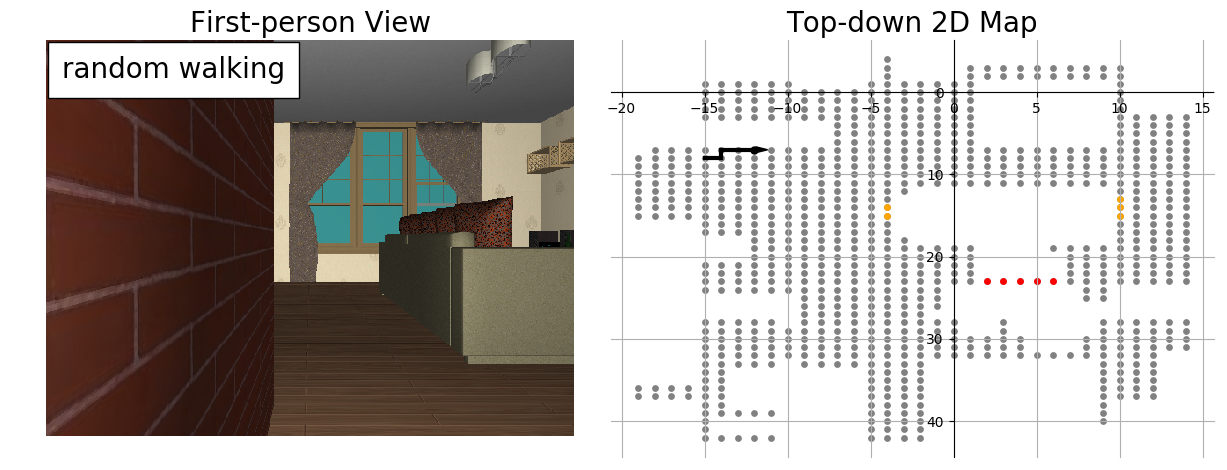} &
\includegraphics[width=0.48\textwidth]{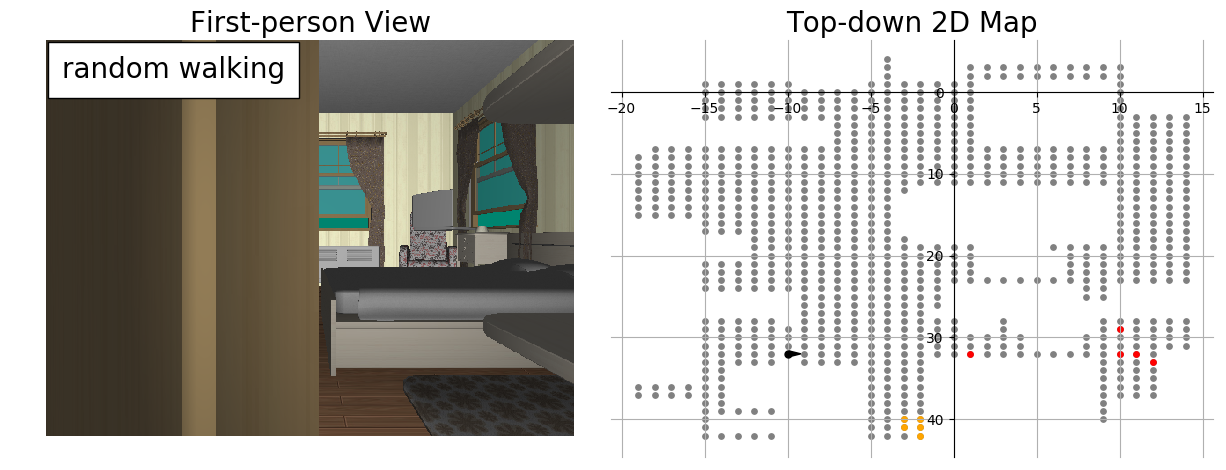} \\
\contour{black}{$\Downarrow$} & \contour{black}{$\Downarrow$} \\
\includegraphics[width=0.48\textwidth]{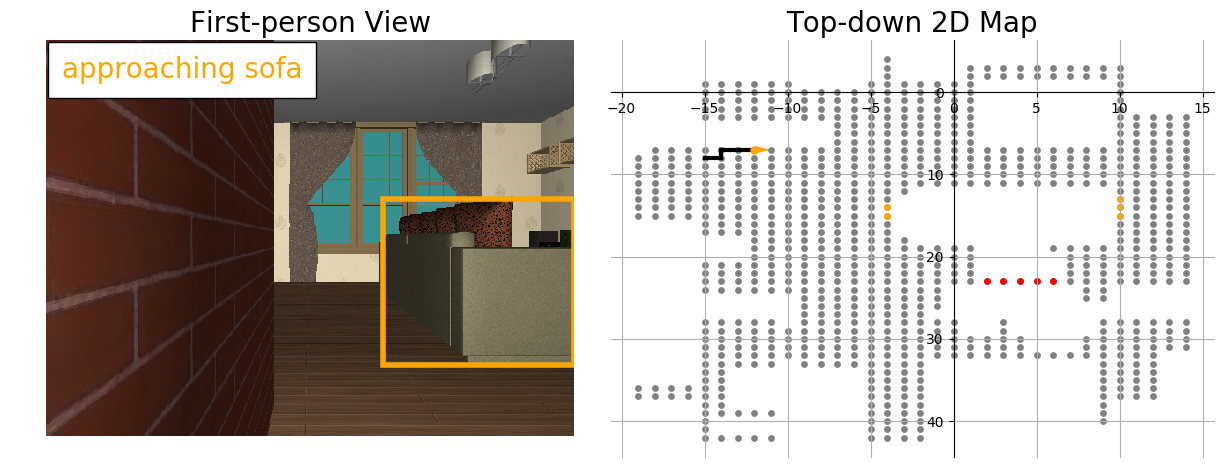} &
\includegraphics[width=0.48\textwidth]{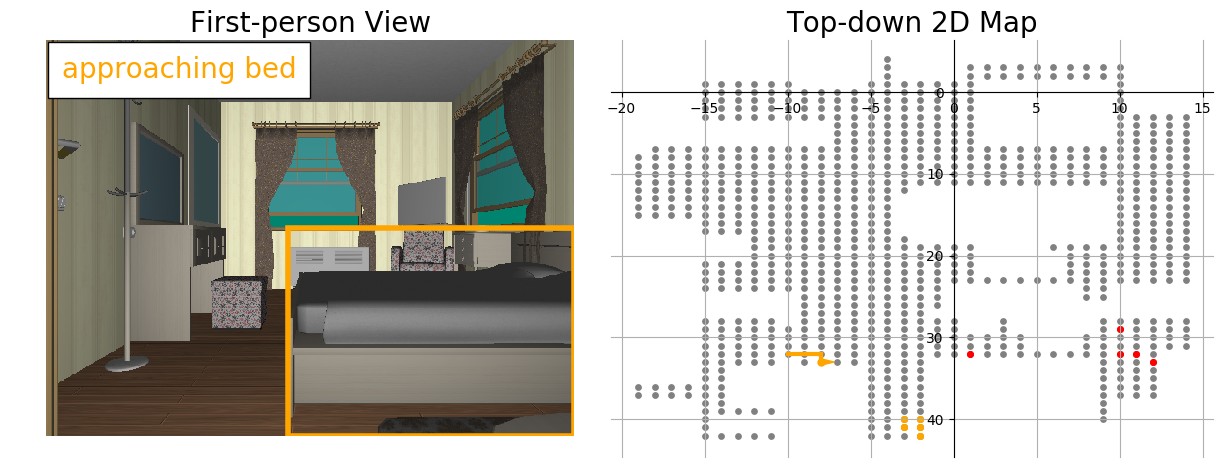} \\
\contour{black}{$\Downarrow$} & \contour{black}{$\Downarrow$} \\
\includegraphics[width=0.48\textwidth]{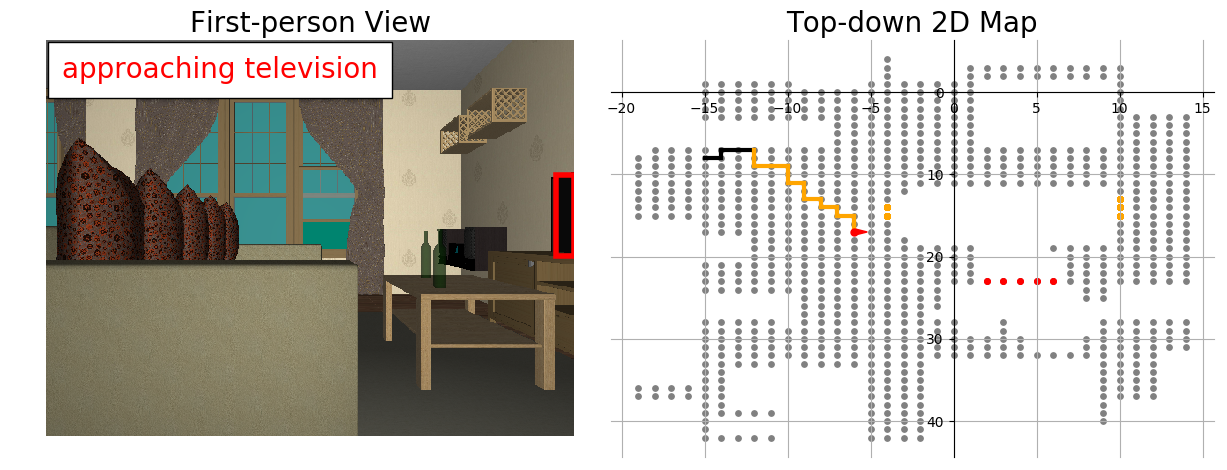} &
\includegraphics[width=0.48\textwidth]{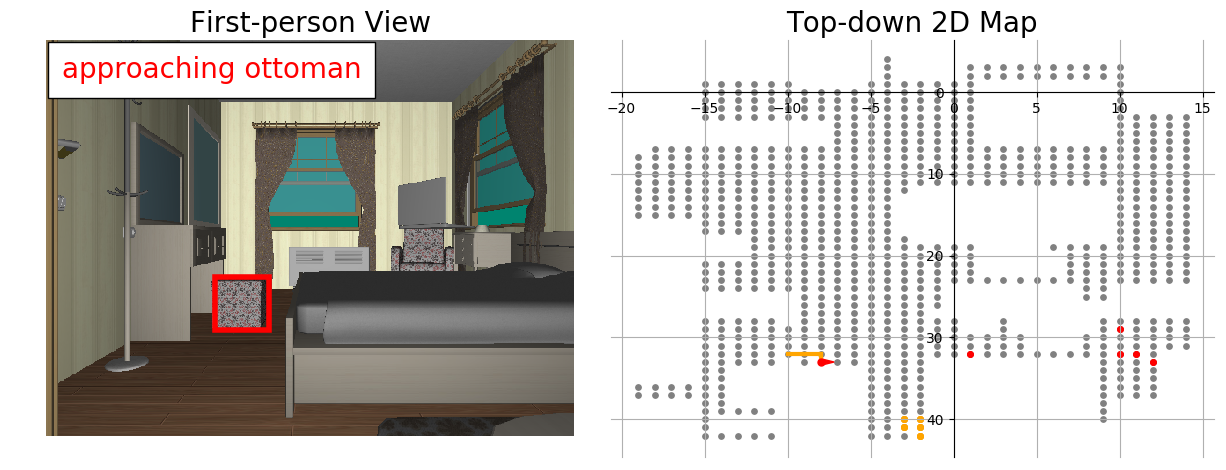} \\
\contour{black}{$\Downarrow$} & \contour{black}{$\Downarrow$} \\
\includegraphics[width=0.48\textwidth]{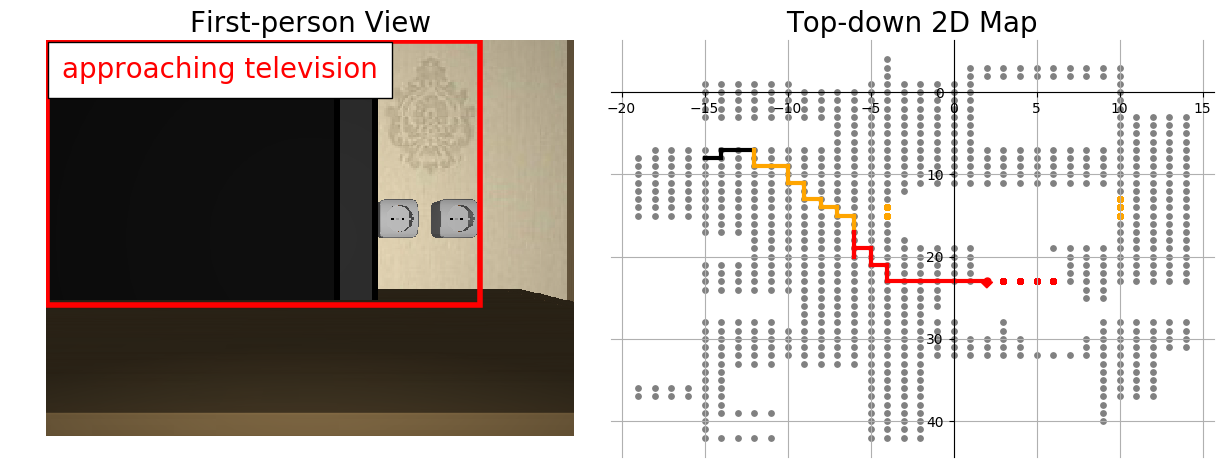} &
\includegraphics[width=0.48\textwidth]{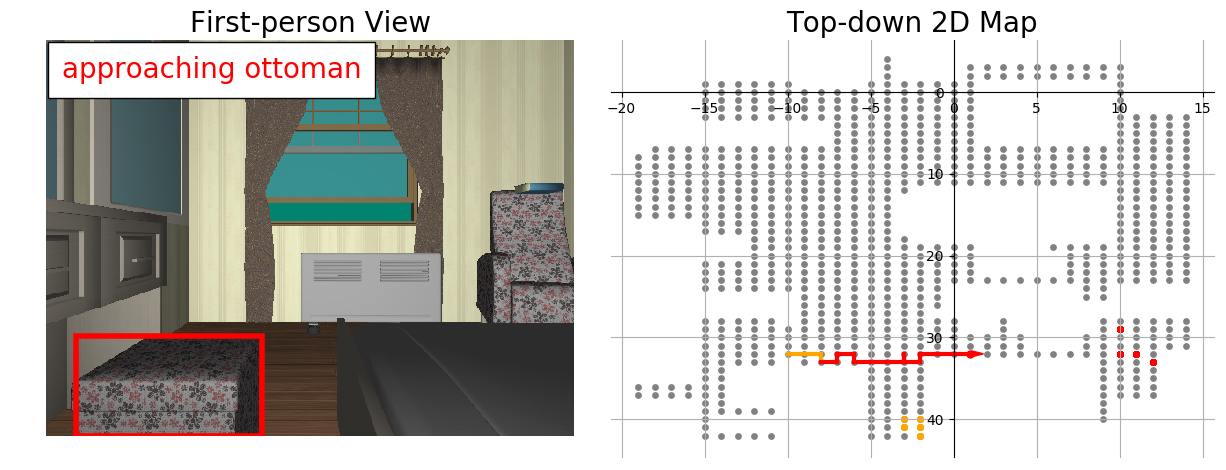} \\
(a) seen environment seen goal & (b) seen environment unseen goal \\
\end{tabular}
% \caption{}
% \label{fig:qua_results}
\end{figure*}

% \ifx\withappendix\undefined
% \small
% \bibliographystyle{ieee_fullname}
% \bibliography{reference}
% \end{document}
% \fi

\end{document}